%% file: main.tex
\documentclass[10pt,twocolumn,letterpaper]{article}

\usepackage{cvpr}              

\input{preamble}

\definecolor{cvprblue}{rgb}{0.21,0.49,0.74}
\usepackage[pagebackref,breaklinks,colorlinks,citecolor=cvprblue]{hyperref}
\usepackage{multirow}
\usepackage{algorithm}
\usepackage{algorithmic}
\usepackage{graphicx}
\usepackage{pifont}
\usepackage{makecell}
\usepackage{amssymb}

\newcommand{\revise}[1]{\color{black}#1\color{black}}

\def\paperID{11545} 
\def\confName{CVPR}
\def\confYear{2024}

\title{Infrared Small Target Detection with Scale and Location Sensitivity}

\author{\revise{Qiankun Liu\textsuperscript{1} \quad Rui Liu\textsuperscript{1,2} \quad Bolun Zheng\textsuperscript{3} \quad Hongkui Wang\textsuperscript{3} \quad Ying Fu\textsuperscript{1,2}\thanks{\revise{Corresponding author}}} \\
\textsuperscript{1}\revise{Beijing Institute of Technology} \quad
\textsuperscript{2}\revise{Yangtze Delta Region Academy of Beijing Institute of Technology} \\
\textsuperscript{3}\revise{Hangzhou Dianzi University}\\
{\tt\small \revise{\{liuqk3, liurui20\}@bit.edu.cn, \{blzheng, wanghk\}@hdu.edu.cn, fuying@bit.edu.cn}}
}

\begin{document}
\maketitle


\begin{abstract}

Recently, infrared small target detection (IRSTD) has been dominated by deep-learning-based methods. However, these methods mainly focus on the design of complex model structures to extract discriminative features, leaving the loss functions for IRSTD under-explored. For example, the widely used Intersection over Union (IoU) and Dice losses lack sensitivity to the scales and locations of targets, limiting the detection performance of detectors. 
In this paper, we focus on boosting detection performance with a more effective loss but a simpler model structure. Specifically, we first propose a novel Scale and Location Sensitive (SLS) loss to handle the limitations of existing losses: 1) for scale sensitivity, we compute a weight for the IoU loss based on target scales to help the detector distinguish targets with different scales: 2) for location sensitivity, we introduce a penalty term based on the center points of targets to help the detector localize targets more precisely. Then, we design a simple Multi-Scale Head to the plain U-Net (MSHNet). By applying SLS loss to each scale of the predictions, our MSHNet outperforms existing state-of-the-art methods by a large margin. In addition, the detection performance of existing detectors can be further improved when trained with our SLS loss, demonstrating the effectiveness and generalization of our SLS loss.  \revise{The code is available at } \url{https://github.com/ying-fu/MSHNet}.

\end{abstract}

\begin{figure}[t]
    \centering
    \includegraphics[width=1.0\linewidth]{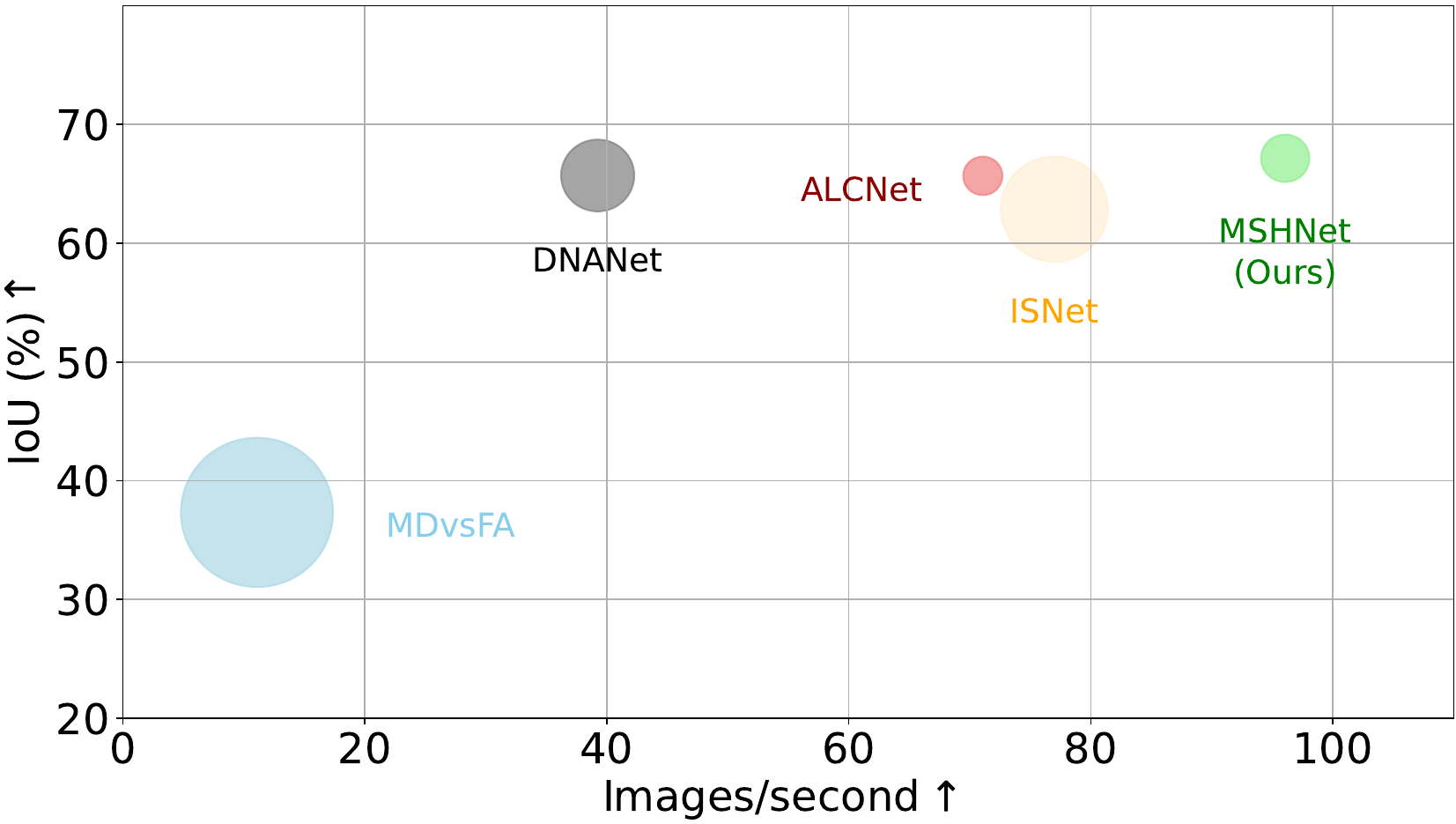}
    \caption{Visualization of the detection performance (IoU), inference time consumption (Images/second) as well as the number of floating point of operations (area of circles) of some deep-learning-based methods. It can be seen that our MSHNet achieves a better balance between these three metrics than other methods. \revise{Results are evaluated on IRSTD-1k \cite{zhang2022isnet}.}}
    \label{fig:motivation_t_iou}
\end{figure}

\section{Introduction}
\label{sec:intro}
Infrared small target detection (IRSTD) is an important computer vision task, which has a wide range of applications,  such as maritime surveillance \cite{maritime_1, field_2}, traffic management \cite{field_1, zhang2022isnet, ying2023mapping} and so on. However, due to the long-distance camera capture, and the noise and clutter interference, infrared targets are often small and dim, making it difficult to detect infrared small targets effectively.

To enhance the detection performance of infrared small targets, numerous methods have been proposed. Early traditional methods can be divided into filtering-based methods \cite{tophat, max_mean}, \revise{local-contrast-based } methods \cite{wslcm, tllcm}, and \revise{low-rank-based } methods \cite{nram, RIPT, PSTNN, zhang2021hyperspectral}. However, these traditional methods rely on manually designed features, and thus can not generalize well to environment changes. Recently, with the development of Deep Learning (DL), IRSTD has been dominated by DL-based methods \cite{liuinfrared, mdvsfa, wei2022tfpnp, chen2023instance}. Different from traditional methods, DL-based methods can automatically learn useful features through a gradient descent algorithm with the constraints of loss functions, making them more robust to various scenarios.

\begin{figure}[t]
	\centering
	\includegraphics[width=0.945\linewidth]{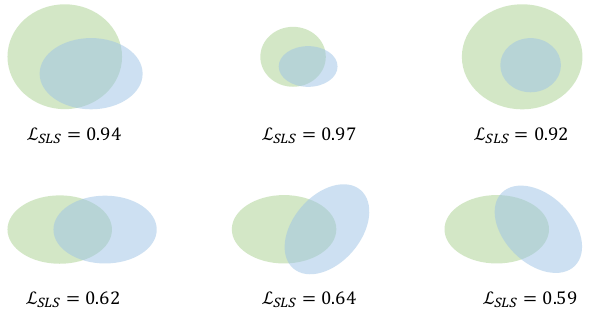}
	\vspace{-5pt}
	\caption{Top row: our SLS loss for the targets of different scales, where IoU loss (=0.4) and Dice loss (=0.57) have the same values for different cases. Bottom row: our SLS loss for the targets of different locations, where IoU loss (=0.3) and Dice loss (=0.43) have the same values for different cases.}
	\label{fig:motivation}
\end{figure}

However, existing DL-based methods primarily focus on designing complex model structures for feature extraction, leaving the loss functions for IRSTD under-explored. For instance, Li \textit{et.al.} \cite{dnanet} customize a dense nested interactive module to achieve multi-layer feature fusion, and Wu \textit{et.al.} \cite{uiunet} nest the U-Net structure to achieve feature aggregation. Although discriminative features can be extracted by the complex model structures, the detection performance is still limited by the under-explored loss functions. For example, the widely used Intersection over Union (IoU) loss and Dice loss \cite{diceloss} lack sensitivity to the scales and locations of targets. As shown in \cref{fig:motivation}, targets with different scales (top row) and locations (bottom row) may share the same IoU loss or Dice loss. This insensitivity to scales and locations makes it challenging for detectors to distinguish targets of different scales and locations, which ultimately limits the detection performance.

In this paper, we focus on boosting detection performance with a more effective loss function but a simpler model structure. Specifically, we first propose a novel Scale and Location Sensitive (SLS) loss to handle the limitations of existing losses.  The merits of the proposed SLS loss include: (1) Scale sensitivity. We compute a weight for the IoU loss based on the predicted and ground-truth scales of targets. The larger the gap between predicted and ground-truth scales is, the more attention will be paid by the detector. (2) Location Sensitivity. We design a location penalty based on the predicted and ground-truth center points of targets. \revise{Compared with traditional L1 and L2 distances, the designed location penalty produces the same value for fewer different location errors}, making the detector locate targets more precisely. Then, we introduce a simple Multi-Scale Head to the plain U-Net (MSHNet), which produces multi-scale predictions for each input. Through leveraging SLS loss at different scales, our MSHNet outperforms existing state-of-the-art (SOTA) methods by a large margin. With the absence of complex structures, our detector achieves a better balance between detection performance, floating point of operations (FLOPs) and inference time consumption, as shown in \cref{fig:motivation_t_iou}.
Moreover, we further train different existing detectors with our SLS loss and achieve better detection performance, demonstrating the effectiveness and generalization of our SLS loss.
 
In summary, our main contributions are:
 \begin{itemize}
    \item We propose a novel scale and location sensitive loss for infrared small target detection, which helps detectors distinguish objects with different scales and locations.
    \item We propose a simple but effective detector by introducing a multi-scale head to the plain U-Net, which achieves SOTA performance without bells and whistles. 
    \item We apply our loss to existing detectors and show that the detection performance can be further boosted, demonstrating the effectiveness and generalization of our loss.
 \end{itemize}

\section{Related Work}
\label{sec:related}
In this section, we first make a brief introduction to existing methods for IRSTD.  Then we provide a review of related works from the perspectives of loss functions and model structures for IRSTD. 

\subsection{Infrared Small Target Detection}
Existing IRSTD methods can be roughly divided into traditional methods and Deep-Learning-based (DL-based) methods. Among them, traditional methods depend on the hand-crafted priors and can be further divided into filtering-based methods \cite{tophat, max_mean}, local-contrast-based methods \cite{wslcm, tllcm, huang2019image}, and low-rank-based methods \cite{ipi, nram, RIPT, fu2020simultaneous, zhang2022deep, PSTNN, MSLSTIPT}. To get useful features with such hand-crafted priors, lots of hyper-parameters need to be manually fine-tuned, making them less robust to the interference of noises and clutters.
Differently, DL-based methods \cite{liuinfrared, mdvsfa, acmnet, fuying-2021-neurocomputing, zhang2022guided, dnanet, zhang2022isnet, uiunet} can automatically learn useful features with the help of loss functions and gradient descent algorithms. Though impressive detection performance has been achieved, existing DL-based methods mainly focus on the design of model structures for the pursuit of more effective features, leaving the loss functions for IRSTD under-explored. 

Different from existing DL-based methods, we focus on boosting detection performance with a more effective loss function but a simpler model structure. A better balance between detection performance, FLOPs and inference time consumption can be achieved.

\subsection{Loss Functions for IRSTD}
As one of the components in DL-based methods, the loss function plays a critical role in the learning process of deep models by quantifying the disparity between predictions and ground-truths. The commonly adopted IoU loss and Dice loss \cite{diceloss} suffer from insensitivity to the scales and locations of targets, rendering the detectors for distinguishing the targets of different scales and locations accurately.

In order to achieve better detection performance, researchers have developed several loss functions. For example, the loss for adversarial training \cite{mdvsfa}, edge loss for the detecting of target edges \cite{zhang2022isnet} and the likelihood loss between target and background maps \cite{likelihood_loss}. However, these losses are tailored for specific network architectures, limiting their broader utility. Different from these dedicated losses, the generalized IoU (GIoU) \cite{giou} and complete IoU (CIoU) \cite{diou} losses have been adopted for box-level IRSTD \cite{irstd_giou,irstd_ciou}. However, these IoU variant losses still lack the sensitivity of scales and locations. 

In contrast to these losses, we formulate a general loss function that is better suitable for IRSTD. It can distinguish targets of different scales and locations, enabling different detectors to achieve better detection performance.

\subsection{Model Structures for IRSTD}
The deep model is another key component in DL-based methods. Liu \textit{et al.} \cite{liuinfrared}  pioneer the application of deep learning to IRSTD. The adopted five-layer Multi-Layer Perceptron (MLP) network demonstrates the superiority of DL-based methods in IRSTD. Recently, lots of works have focused on the design of model structures to get more effective features. Li \textit{et al.} \cite{dnanet} customize a dense nested interactive module to achieve multi-layer feature fusion. Zhang \textit{et al.} \cite{zhang2022isnet} utilize Taylor finite difference and orientation attention strategy to extract edge information of targets. However, such complex model structures not only bring more computational cost but still suffer from moderate detection performance due to the lack of effective loss functions. 

Differently we introduce a simple multi-scale head to the plain U-Net, rather than designing complex structures. By applying our SLS loss to different scales, SOTA performance is achieved with less time consumption.

\begin{figure}
    \centering
    \includegraphics[width=1.0\linewidth]{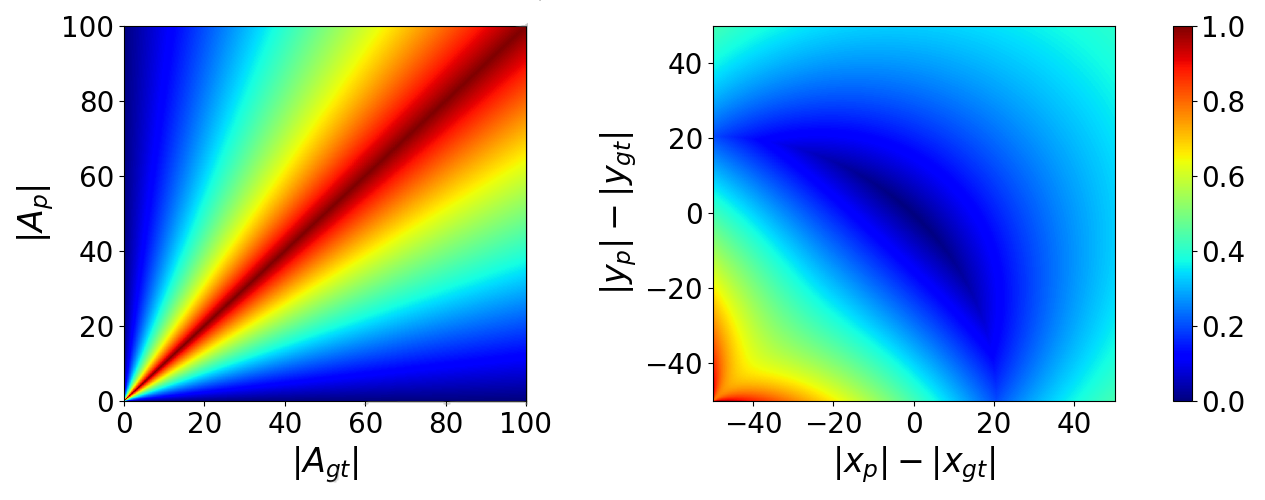}
    \caption{Left: the value of weight $w$ in scale sensitive loss with respect to the number of predicted pixels and ground-truth pixels (\ie, predicted and ground-truth scales). Right: the normalized location sensitive loss with respect to the location error between the predicted center point and ground-truth center point. The range of [0, 100] pixels is shown for illustration.}
    \label{fig:scale_loss_weight_and_location_loss}
\end{figure}

\begin{figure*}[t]
    \centering
    \includegraphics[width=\linewidth]{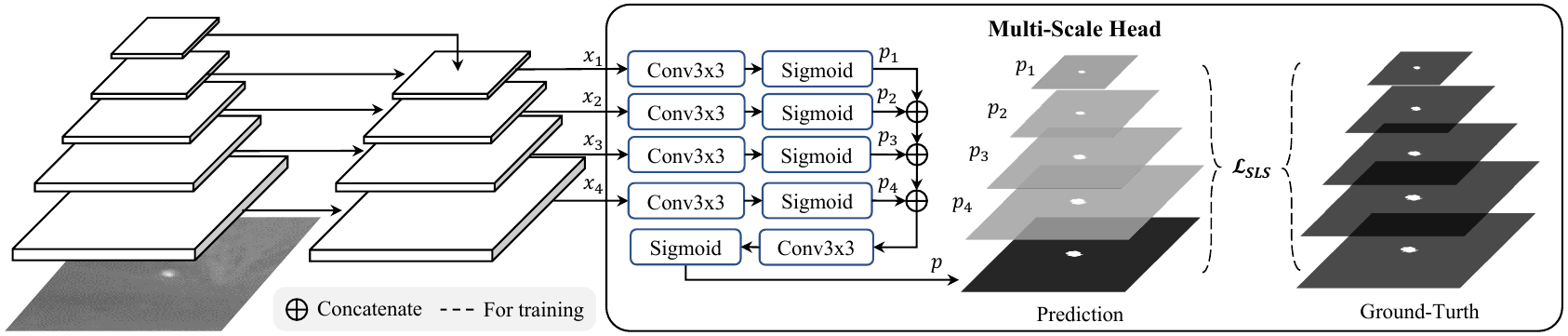}
    \caption{Overview of the proposed MSHNet. Our MSHNet is implemented based on a plain U-Net without bells and whistles. Only a simple multi-scale head is introduced. For each scale, the feature map is fed into a dedicated head, producing a prediction with the same spatial shape as the feature map. Different scales of predictions are \revise{upsampled (if needed) } and concatenated together to get the final prediction. In the training stage, our SLS loss is applied to each of these predictions since it is scale sensitive.}
    \label{fig:network}
\end{figure*}

\section{Scale and Location Sensitive Loss}
\label{sec:sls_loss}
The scale and location sensitive (SLS) loss, denoted as $\mathcal{L}_{SLS}$, is designed to handle the insensitivity of scales and locations in existing losses. It consists of a scale sensitive loss and a location sensitive loss. Formally, 
\begin{equation}
	\mathcal{L}_{SLS} = \mathcal{L}_{S} +\mathcal{L}_{L},
\end{equation}
where $ \mathcal{L}_{S}$ and $\mathcal{L}_{L}$ present the scale sensitive loss and location sensitive loss, respectively.  In the following, we introduce the SLS loss in detail starting from the scale sensitive loss,  which is based on the commonly used IoU loss.

\subsection{Scale Sensitive Loss}
\label{sec:scale_sensitive_loss}
Let $A_{p}$ and $A_{gt}$ be the set of predicted pixels and ground-truth pixels of targets, the IoU loss between them can be formulated as:
\begin{equation}
	\mathcal{L}_{IoU} = 1 - \frac{ |A_{p} \cap A_{gt} | }{ | A_{p} \cup A_{gt} | }.
\end{equation}

Though having been widely used in IRSTD, the IoU loss is insensitive to the scales and locations of targets, as shown in \cref{fig:motivation}. The scale sensitive loss is implemented by providing a weight to the IoU loss:  
\begin{equation}
	\begin{split}
		&\mathcal{L}_{S} = 1 - w \frac{ |A_{p} \cap A_{gt} | }{ | A_{p} \cup A_{gt} | }, \\
		{\rm s.t.} \quad w = &\frac{ {\rm min} (|A_{p}|, |A_{gt}|) +  {\rm Var}(|A_{p}|, |A_{gt}|)} {{\rm max}  (|A_{p}|, |A_{gt}|) + {\rm Var}(|A_{p}|, |A_{gt}|)},
	\end{split}
 \label{eq:loss_s}
\end{equation}
where ${\rm Var(\cdot, \cdot)}$ is the function that gets the variance of provided scalars. 

On the left of \cref{fig:scale_loss_weight_and_location_loss}, we visualize the values of $w$ with respect to the number of pixels in \revise{$A_p$ and $A_{gt}$}. It can be observed that the larger the gap between $|A_{p}|$ and $|{A_{gt}}|$ is, the smaller the $w$ is, which results in a larger scale sensitive loss (on the assumption that the IoU between $A_{p}$ and $A_{gt}$ is fixed). The intuition behind the design of $w$ is that the detector should pay more attention to the target with a larger loss if the predicted and ground-truth scales (\ie, the numbers of pixels in $A_{p}$ and $A_{gt}$) are quite different.

\subsection{Location Sensitive Loss}
\label{sec:location_loss}
The location sensitive loss is calculated based on the predicted and ground-truth center points of targets. 
Given the sets of predicted pixels $A_{p}$ and ground-truth pixels $A_{gt}$, the corresponding center points for $A_{p}$ and $A_{gt}$ are obtained by averaging the coordinates of all pixels, which are denoted as $\mathbf{c}_{p} = (x_{p}, y_{p})$  and  $\mathbf{c}_{gt} = (x_{gt}, y_{gt}$), respectively.  Then, we convert the coordinates of these two center points into the polar coordinate system. Take $\mathbf{c}_{p}$ for example, the corresponding distance $d_{p}$ and angle $\theta_{p}$ in the polar coordinate system are:
\begin{equation}
	\begin{split}
		d_p &= \sqrt{x_{p}^{2}+y_{p}^{2}}, \\
		\theta_{p} &= \arctan(\frac{y_{p}}{x_{p}}).
	\end{split}
\end{equation}
The location sensitive loss can be obtained by:
\begin{equation}
	\mathcal{L}_{L} = (1 - \frac{{\min( d_{p}, d_{gt} )}} {\max ( d_{p}, d_{gt})}) + \frac{4}{\pi^2}(\theta_{p} - \theta_{gt})^2 ,
\end{equation}
where $d_{gt}$ and $\theta_{gt}$ are the distance and angle of $\mathbf{c}_{gt}$ in the polar coordinate system, respectively. 

On the right of \cref{fig:scale_loss_weight_and_location_loss}, we show how the location loss changes with respect to different location errors between $\mathbf{c}_{p}$ and $\mathbf{c}_{gt}$. As we can see,  though some different location errors share the same loss value \revise{(these location errors still can be distinguished with their gradients)}, the location loss distinguishes most of the different location errors effectively, making the detector sensitive to different types of location errors and locate the targets more accurately.

\section{MSHNet Dectector}
\label{sec:mshnet}
In this section, we introduce our MSHNet detector, which is implemented by introducing a simple but effective multi-scale head to the plain U-Net. The overview of MSHNet is shown in \cref{fig:network}. We take the commonly used U-Net as the backbone network. The feature maps that have different scales in the decoder are fed into different prediction heads to get different scales of predictions. All the predictions from different feature maps are finally concatenated (upsampling before concatenation is adopted if needed) to get the final prediction. In the training stage, our SLS loss is applied to each of the predictions. In the following, the multi-scale head is first described in detail. Then we introduce the utilization of SLS loss in MSHNet.
 
\subsection{Multi-Scale Head}
\label{sec:multi_scale_head}
Let $\mathbf{x}_i \in \mathbb{R}^{H_i \times W_i \times C_i}$ be the feature map at the $i$-th scale in the decoder of U-Net, where $H_i \times W_i$ is the spatial size and $C_i$ is the number of channels. Following the common settings within existing works, there are 4 scales in U-Net, which means that $i \in \{1,2,3,4\}$. Supposing the input spatial size is $H \times W$, then $H_i=\frac{H}{2^{4-i}}$ and  $W_i=\frac{W}{2^{4-i}}$. 

The $i$-th prediction $\mathbf{p}_i \in \mathbb{R}^{H_i \times W_i \times 1}$ is obtained by the corresponding prediction head, which is implemented by a convolution layer and a sigmoid activation function:
\begin{equation}
\mathbf{p}_i = {\rm Sigmoid}( {\rm Conv}(\mathbf{x}_i)).
\end{equation}
Note that different prediction heads have their own dedicated parameters. The final prediction $\mathbf{p} \in \mathbb{R}^{H \times W \times 1}$ is obtained based on all the 4 predictions:
\begin{equation}
	\mathbf{p} = {\rm Sigmoid}( {\rm Conv}([ \Uparrow (\mathbf{p}_1, 8), \Uparrow (\mathbf{p}_2, 4), \Uparrow (\mathbf{p}_3, 2), \mathbf{p}_4])),
\end{equation}
where $\Uparrow(\cdot, \cdot)$ is the operation that spatially upsamples the first argument with the second argument as the factor, and $[\cdot, ..., \cdot]$ concatenates all the provided arguments along the \revise{channel } dimensionality.

\begin{table*}[t]
	\setlength{\tabcolsep}{16pt}
	\begin{center}
        \small
		\begin{tabular}{c|c|c|c|c|c|c|c}
			\hline
			\multirow{2}{*}{Method} & \multirow{2}{*}{ Description} &\multicolumn{3}{c|}{IRSTD-1k} & \multicolumn{3}{c}{\revise{NUDT-SIRST}}\\
			\cline{3-8}
			& & IoU$\uparrow$ & ${\rm P_d}\uparrow$ & ${\rm F_a}\downarrow$ & IoU$\uparrow$ & ${\rm P_d}\uparrow$ & ${\rm F_a}\downarrow$\\
			\hline
			Top-Hat \cite{tophat} & \multirow{2}{*}{Filtering} & 10.06 & 75.11 & 1432 & 20.72 & 78.41 & 166.7\\
			Max-Median \cite{max_mean} &  & 6.998 & 65.21 & 59.73 & 4.197 & 58.41 & 36.89\\
			\hline
			WSLCM \cite{wslcm} & \multirow{2}{*}{Local Contrast} & 3.452 & 72.44 & 6619 & 2.283 & 56.82 & 1309\\
			TLLCM \cite{tllcm} &  & 3.311 & 77.39 & 6738 & 2.176 & 62.01 & 1608\\
			\hline
			IPI \cite{ipi} & \multirow{5}{*}{Low Rank} & 27.92 & 81.37 & 16.18 & 17.76 & 74.49 & 41.23\\
			NRAM \cite{nram} & & 15.25 & 70.68 & 16.93 & 6.927 & 56.40 & 19.27\\
			RIPT \cite{RIPT} &  & 14.11 & 77.55 & 28.31 & 29.44 & 91.85 & 344.3\\
			PSTNN \cite{PSTNN} & & 24.57 & 71.99 & 35.26 & 14.85 & 66.13 & 44.17\\
			MSLSTIPT \cite{MSLSTIPT} &  & 11.43 & 79.03 & 1524 & 8.342 & 47.40 & 888.1\\
			\hline
			MDvsFA \cite{mdvsfa} & \multirow{5}{*}{Deep Learning} & 37.34 & 83.71 & 88.52 & 35.86 & 85.22 & 95.37\\
			ALCNet \cite{alcnet} & & 65.68 & 89.25 & 27.71 & 72.89 & 96.19 & 30.40\\
			ISNet \cite{zhang2022isnet} & & 62.88 & 92.59 & 27.92 & 67.86 & 92.59 & 34.65\\
			DNANet \cite{dnanet} &  & 65.71 & 91.84 & 17.61 & 79.98 & 96.93 & 12.78\\
			MSHNet (Ours) &  & \textbf{67.16} & \textbf{93.88} & \textbf{15.03} & \textbf{80.55} & \textbf{97.99} & \textbf{11.77}\\
			\hline
		\end{tabular}

	\caption{\revise{Quantitative results of different methods. Results for the metrics of IoU(\%), ${\rm P_d}$(\%) and ${\rm F_a}$($10^{-6}$) are presented. The best values are highlighted with \textbf{bold}. It can be seen that our MSHNet achieves the best results on different metrics and datasets.}}
	\label{tab:main_quantitative_result}
	\end{center}
\end{table*}

\subsection{Training MSHNet with SLS Loss}
Since our SLS loss is scale sensitive and there are several scales in the predictions of MSHNet, we apply SLS loss to all of the predictions. 
The inspiration is that our SLS loss produces different loss values for different scales even if they share the same spatial layout (refer to the first two cases in the top row,  \cref{fig:motivation} \revise{)} . 
We hypothesize that by applying our SLS loss to different scales, the targets that are with different scales can attract different attention from the detector, resulting in an overall better detection performance. 

Let $\mathbf{p}_{gt} \in \{0,1\}^{H\times W \times 1}$ be the ground-truth label. The final loss for MSHNet is:
\begin{equation}
	\mathcal{L} = \frac{1}{5} (\sum_{i=1}^{4}\mathcal{L}_{SLS} (\mathbf{p}_i, \Downarrow(\mathbf{p}_{gt}, 2^{4-i})) + \mathcal{L}_{SLS} (\mathbf{p}, \mathbf{p}_{gt}) ) ,
\end{equation}
where $\Downarrow(\cdot, \cdot)$ is the operation (\ie, max-pooling) that spatially downsamples the first argument with the second argument as the factor, and $\mathcal{L}_{SLS}(\cdot, \cdot)$ is our SLS loss.

\section{Experiments}
\label{sec:experiments}

In this section, the \revise{adopted } datasets and metrics are firstly introduced, followed by the implementation details. Then, we compare the proposed method MSHNet with existing IRSTD detectors. Finally, discussions are provided to show the effectiveness of our Scale and Location Sensitive loss and MHSNet detector.

\subsection{Datasets and Metrics}
\paragraph{Datasets.} 
The experiments are conducted on two datasets \ie, IRSD-1k \cite{zhang2022isnet} and NUDT-SIRST \cite{dnanet}. There are 1,001 and 1,327 infrared images in IRSTD-1k and NUDT-SIRST, respectively. Following existing works \cite{zhang2022isnet,dnanet}, the images in IRSTD-1k are divided into the training and testing splits with a ratio of 4:1, while the images in NUDT-SIRST are equally divided into the training and testing splits.

\paragraph{Evaluation Metrics.}
\revise{We use IoU and false alarm rate ($\rm F_{a}$) as the pixel-level evaluation metric and assess target-level performance using the probability of detection (${\rm P_d}$). Different metrics reveal the performance of the detector from different aspects. $\rm F_a$ and $\rm P_d$ focus on recall and false alarms, while IoU takes both into consideration. }

 Formally, The false alarm rate is:
\begin{equation}
	{\rm F_a} = \frac{P_{false}}{P_{all}},
\end{equation}
where $P_{false}$ is the number \revise{of } false positive pixels and $P_{all}$ is the number of all pixels in the image.

The probability of detection is:
\begin{equation}
     {\rm P_d} = \frac{N_{pred}}{N_{all}},
\end{equation}
where $N_{pred}$ is \revise{the } number of correctly predicted targets and $N_{all}$ is the number of all targets.

\subsection{Implementation Details}
The proposed method is implemented with PyTorch framework. Following existing works, the input size of the detector is set to $256\times 256$. We train different models using AdaGrad on 2 RTX3090 GPUs. The batch size is set to 4 and the learning rate is set to 0.05.

\subsection{Comparison to Existing Methods}
We first compare the proposed MSHNet with existing methods. Different types of methods are evaluated, including traditional methods and deep-learning-based methods. The evaluated traditional methods are the filtering-based Top-Hat \cite{tophat} and Max-Median \cite{max_mean}, local-contrast-based WSLCM \cite{wslcm} and TLLCM \cite{tllcm}, \revise{low-rank-based } IPI \cite{ipi}, NRAM \cite{nram}, RIPT \cite{RIPT}, PSTNN \cite{PSTNN} and MSLSTIPT \cite{MSLSTIPT}.
The deep-learning-based methods are MDvsFA \cite{mdvsfa}, ALCNet \cite{alcnet}, ISNet \cite{zhang2022isnet}, and DNANet \cite{dnanet}. For a fair comparison, all the evaluated deep-learning-based models are retrained \revise{with their official codes } to their convergence on NUDT-SIRST \cite{dnanet} and IRSTD-1k \cite{zhang2022isnet} datasets.

\begin{figure*}
    \centering
    \setlength\tabcolsep{2pt}
    \footnotesize
    \begin{tabular}{cccccccc}
       \makebox[0.12\textwidth][c]{Infrared Images} &  \makebox[0.12\textwidth][c]{Ground-Truth}  & \makebox[0.12\textwidth][c]{Top-Hat\cite{tophat}}  &  \makebox[0.11\textwidth][c]{WSLCM\cite{wslcm} } & 
       \makebox[0.12\textwidth][c]{IPI\cite{ipi}} & 
       \makebox[0.12\textwidth][c]{ISNet\cite{zhang2022isnet}} & \makebox[0.12\textwidth][c]{DNANet\cite{dnanet}}  &  \makebox[0.1\textwidth][c]{MSHNet (Ours)} \\
        
        \multicolumn{8}{c}{\includegraphics[width=\textwidth]{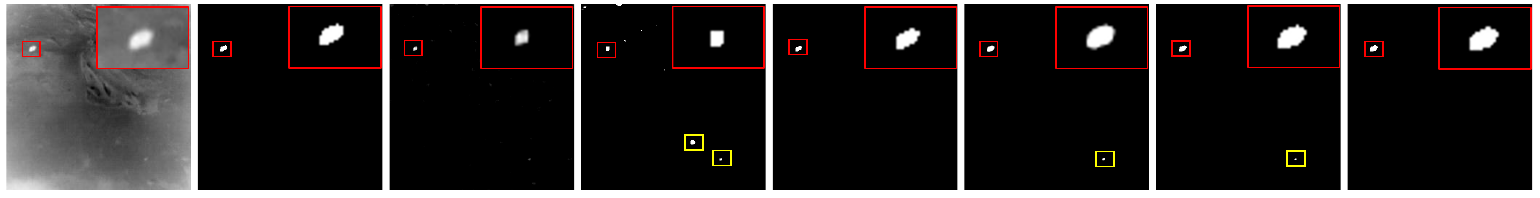}} \\
        \multicolumn{8}{c}{\includegraphics[width=\textwidth]{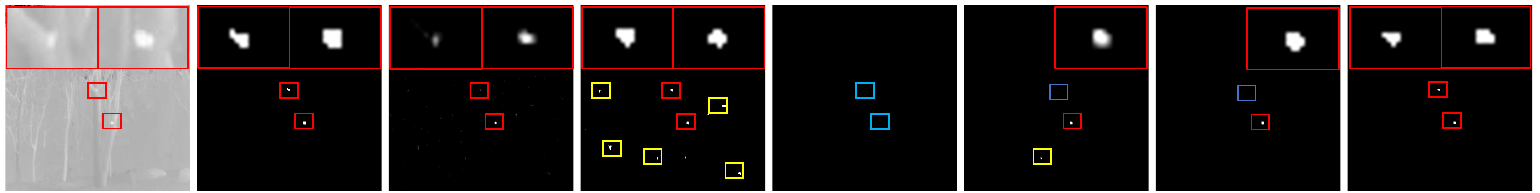}} \\
        \multicolumn{8}{c}{\includegraphics[width=\textwidth]{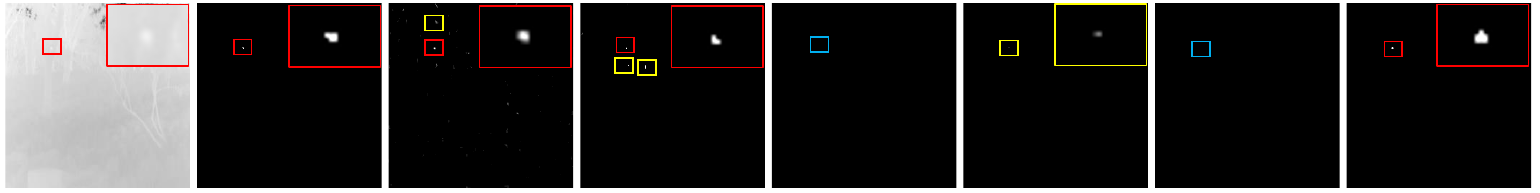}} \\ 
        \multicolumn{8}{c}{\includegraphics[width=\textwidth]{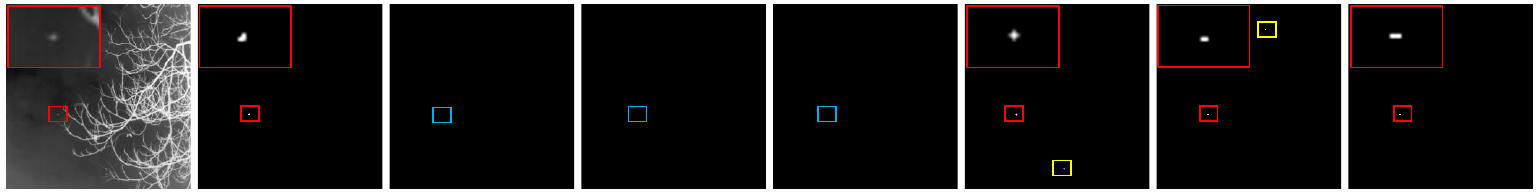}} \\
        \multicolumn{8}{c}{\includegraphics[width=\textwidth]{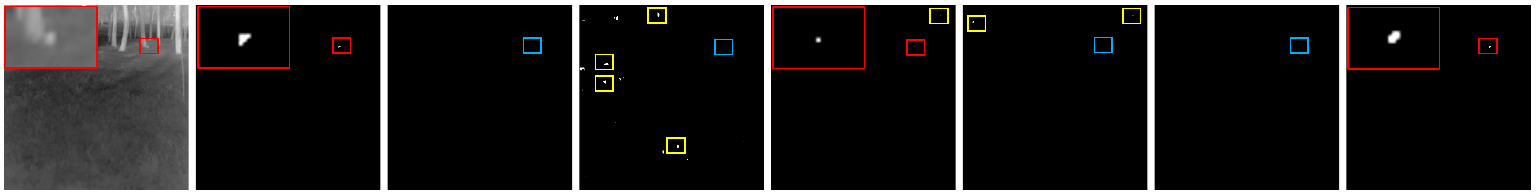}} \\
    \end{tabular}
 \caption{Visual comparison of detection results on several infrared images. 
      	Correctly detected targets, missed targets, and false alarms are framed by red, blue, and yellow boxes, respectively. A close-up view of the target is shown in image corners. }
     \label{fig:results_visualization}
\end{figure*}

\begin{figure*}
    \centering
    \setlength\tabcolsep{2pt}
    \footnotesize
    \begin{tabular}{cccccccc}
       \makecell[c]{Infrared Images} &   \makecell[c]{Ground-Truth}  &  \makecell[c]{Top-Hat\cite{tophat}}  &  \makecell[c]{WSLCM\cite{wslcm}}  &  \makecell[c]{IPI\cite{ipi}} &  \makecell[c]{ISNet\cite{zhang2022isnet}}  &  \makecell[c]{DNANet\cite{dnanet}}  &  \makecell[c]{MSHNet (Ours)} \\
        \includegraphics[width=0.115\textwidth]{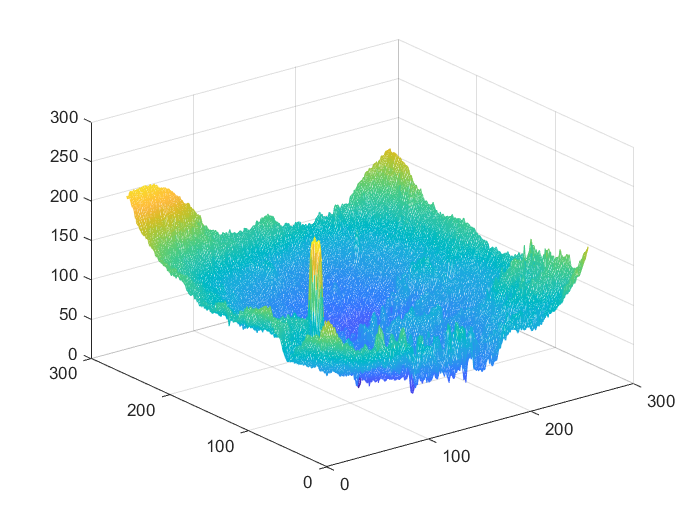} & \includegraphics[width=0.115\textwidth]{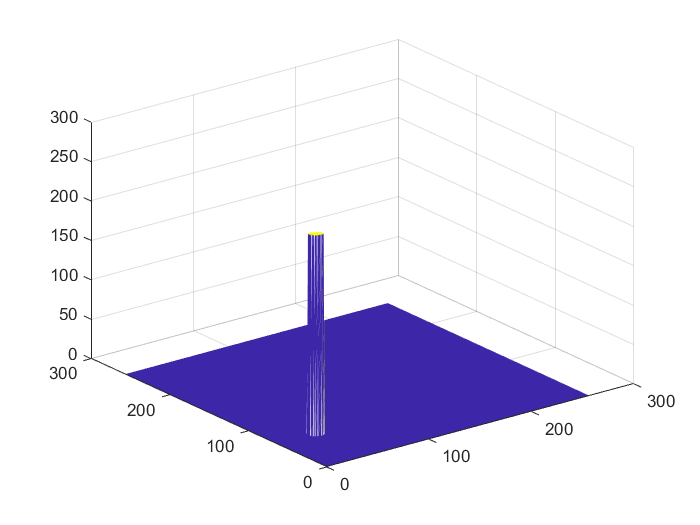} & \includegraphics[width=0.115\textwidth]{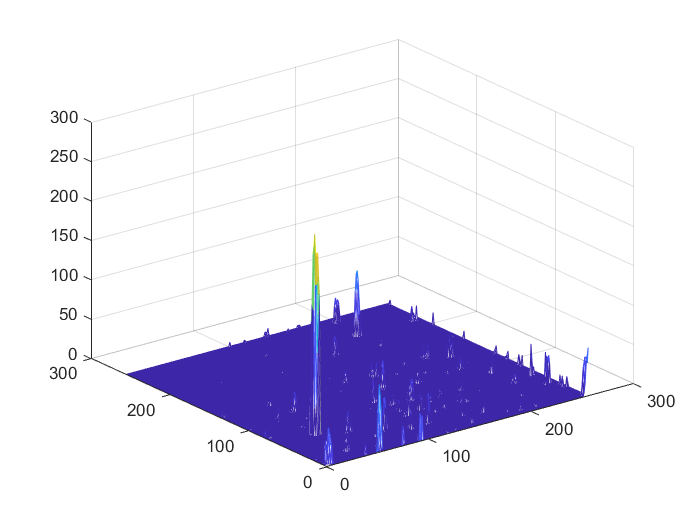} & \includegraphics[width=0.115\textwidth]{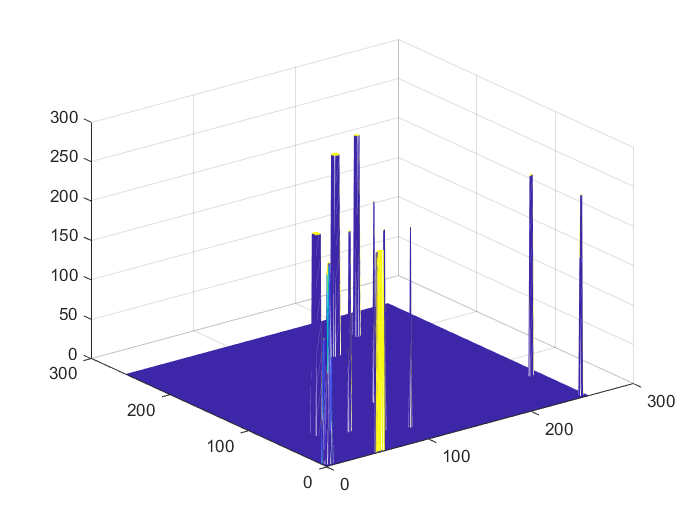} & \includegraphics[width=0.115\textwidth]{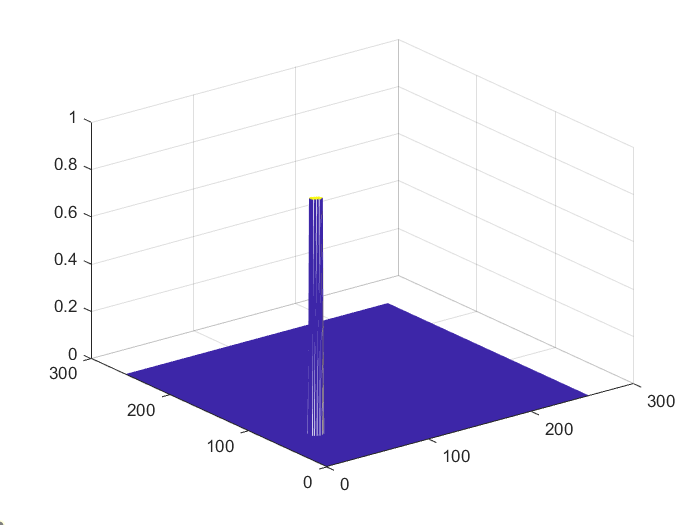} & \includegraphics[width=0.115\textwidth]{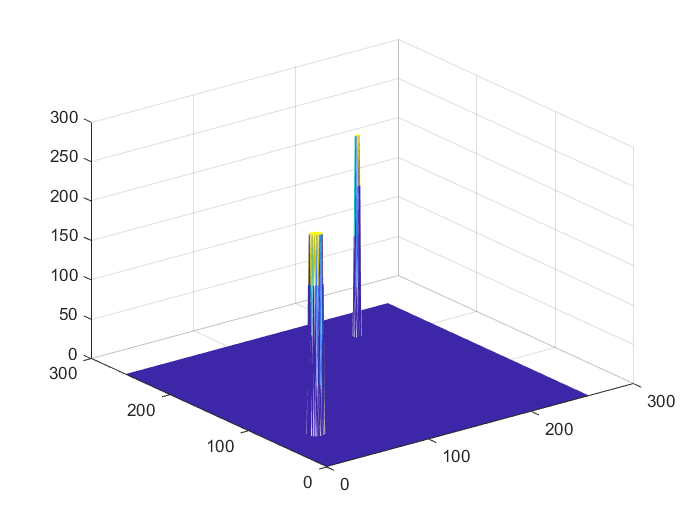} & \includegraphics[width=0.115\textwidth]{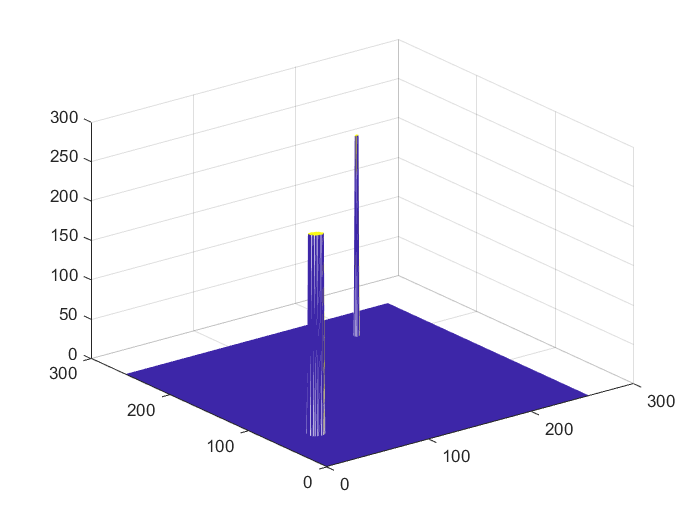} & \includegraphics[width=0.115\textwidth]{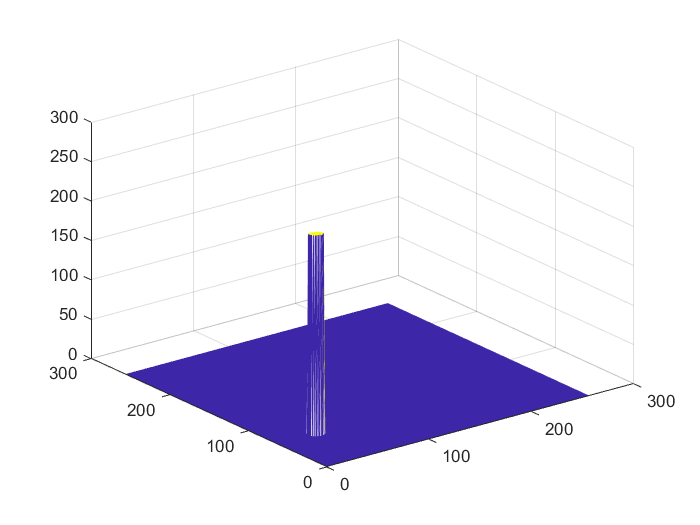} \\
        
        \includegraphics[width=0.115\textwidth]{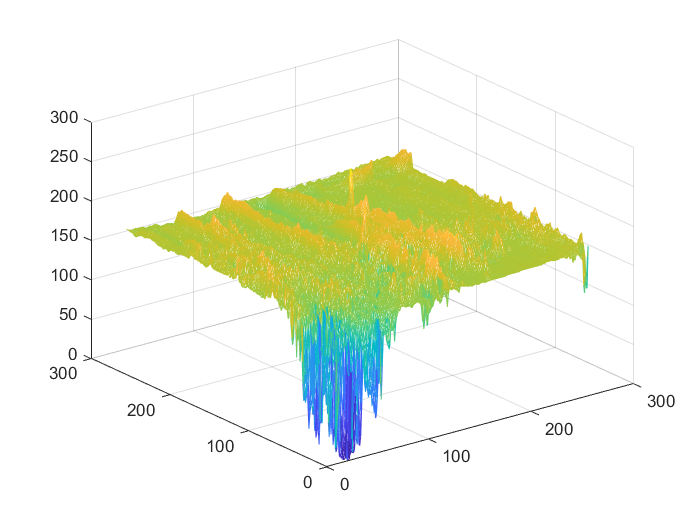} & \includegraphics[width=0.115\textwidth]{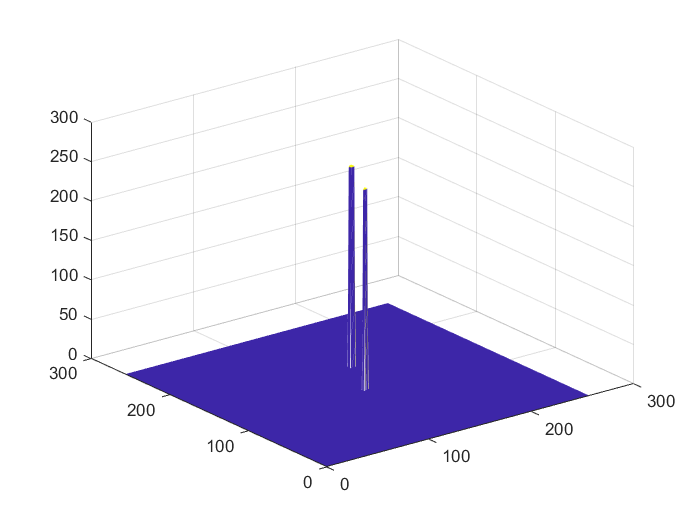} & \includegraphics[width=0.115\textwidth]{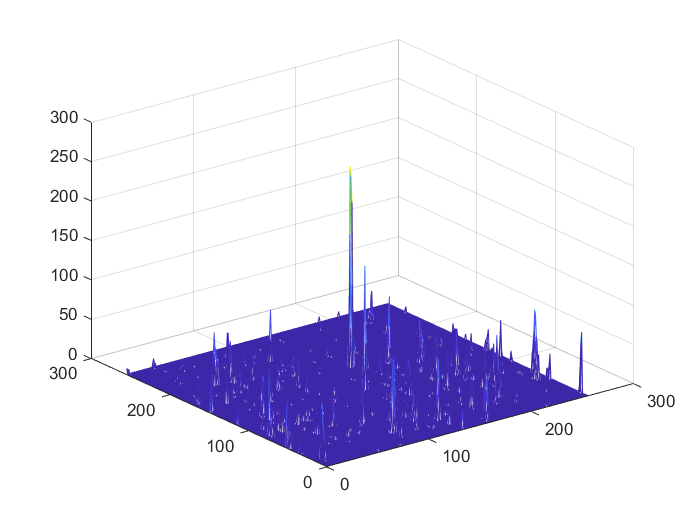} & \includegraphics[width=0.115\textwidth]{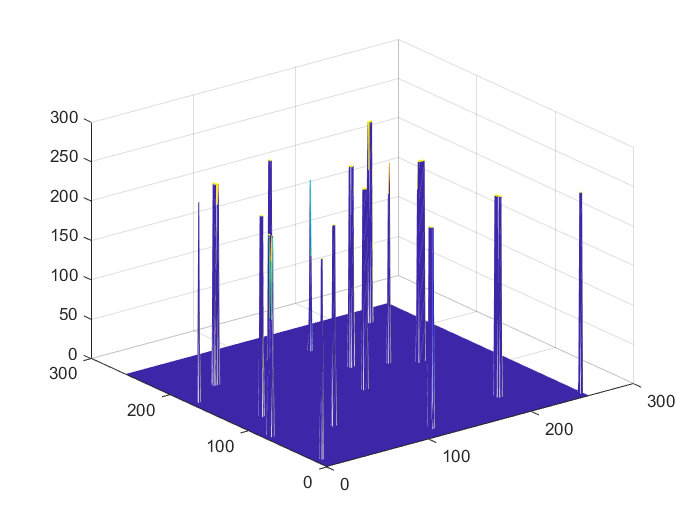} & \includegraphics[width=0.115\textwidth]{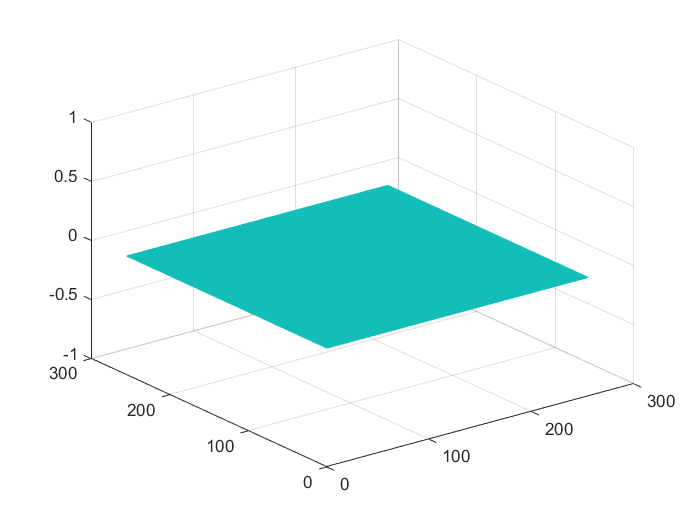} & \includegraphics[width=0.115\textwidth]{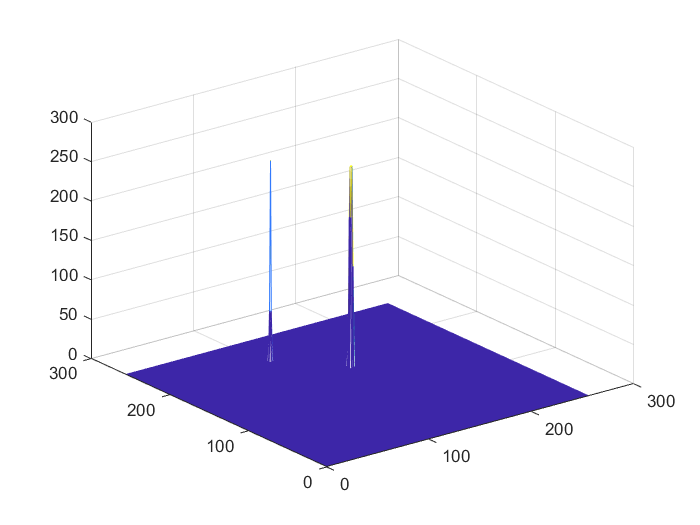} & \includegraphics[width=0.115\textwidth]{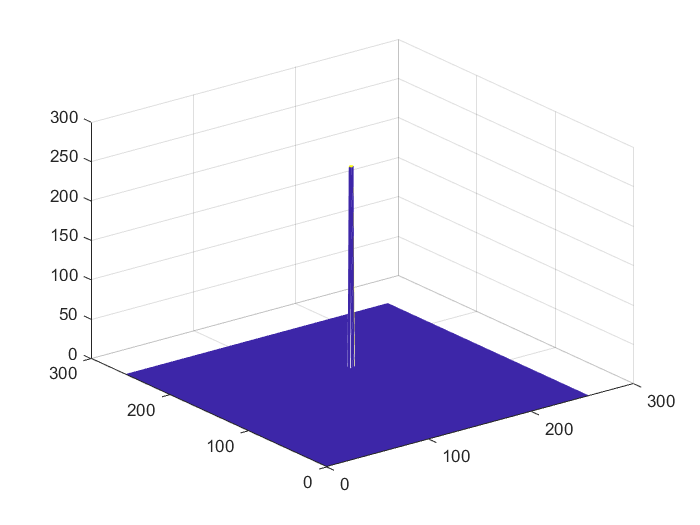} & \includegraphics[width=0.115\textwidth]{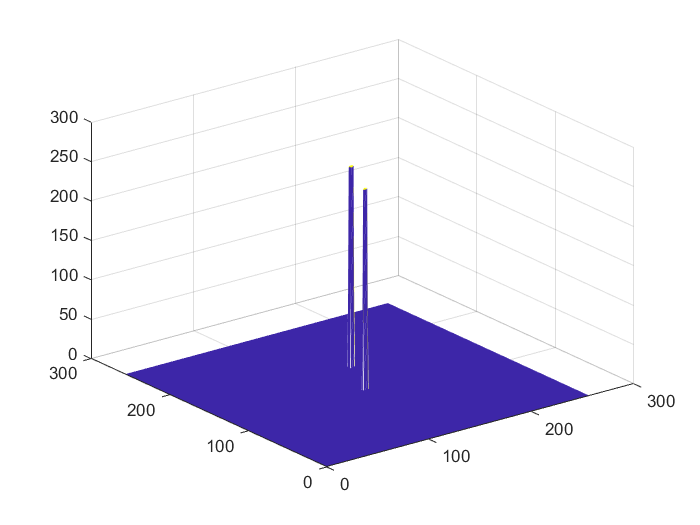} \\

        \includegraphics[width=0.115\textwidth]{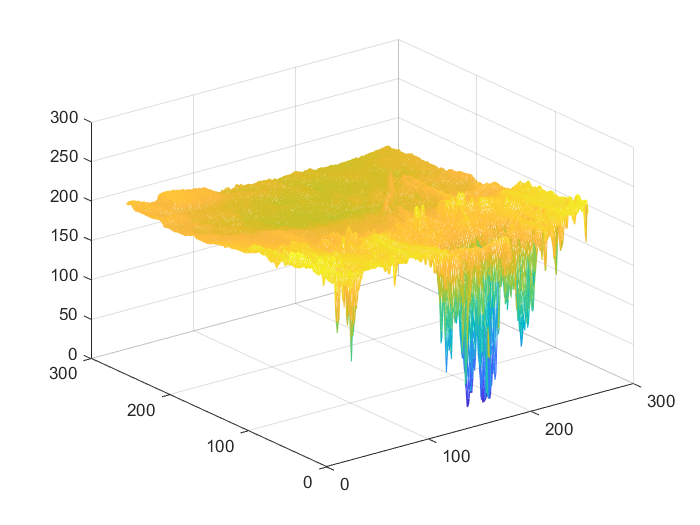} & \includegraphics[width=0.115\textwidth]{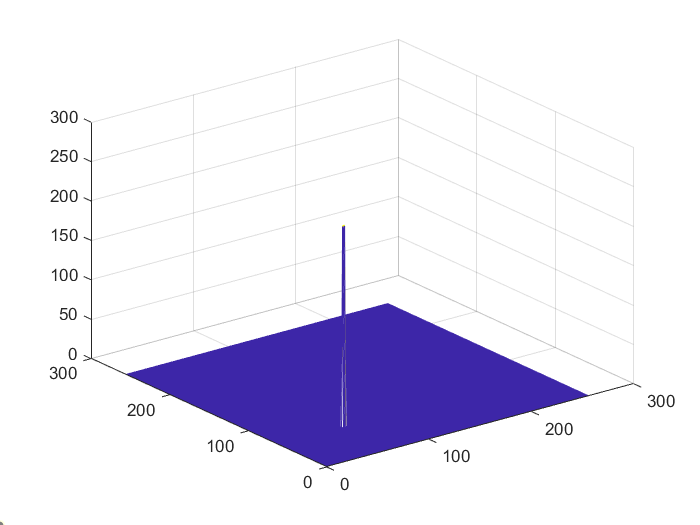} & \includegraphics[width=0.115\textwidth]{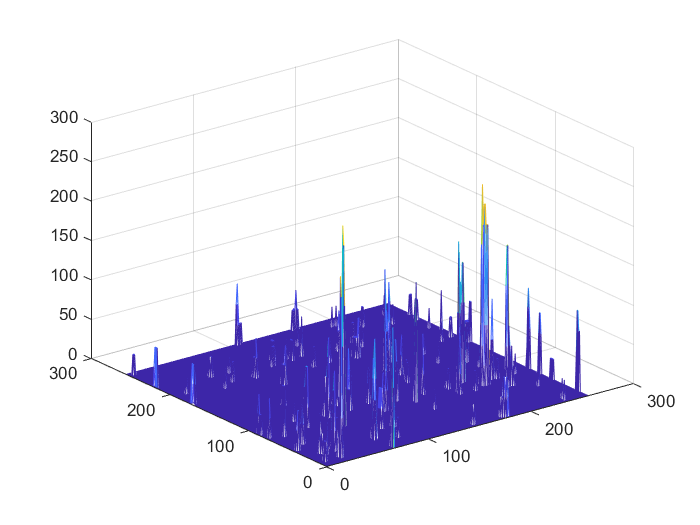} & \includegraphics[width=0.115\textwidth]{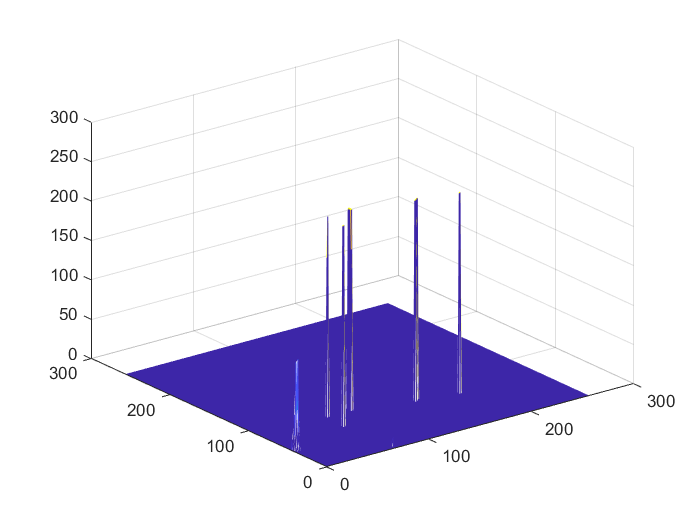} & \includegraphics[width=0.115\textwidth]{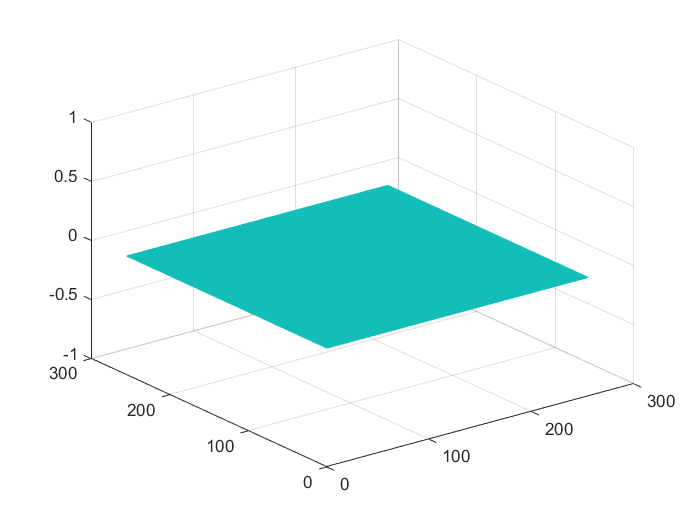} & \includegraphics[width=0.115\textwidth]{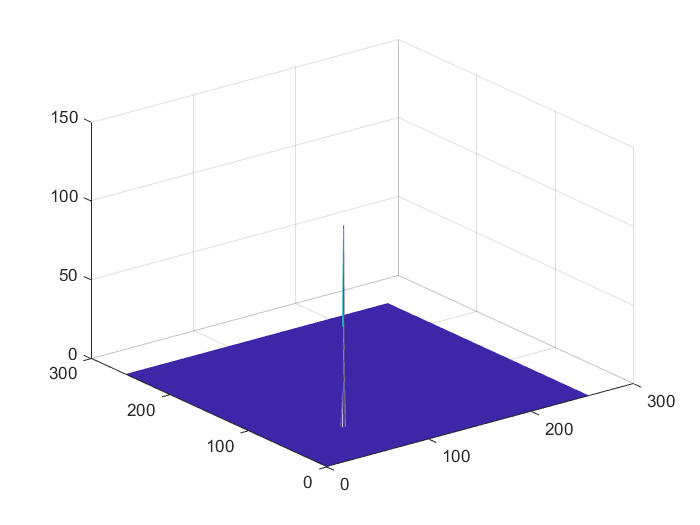} & \includegraphics[width=0.115\textwidth]{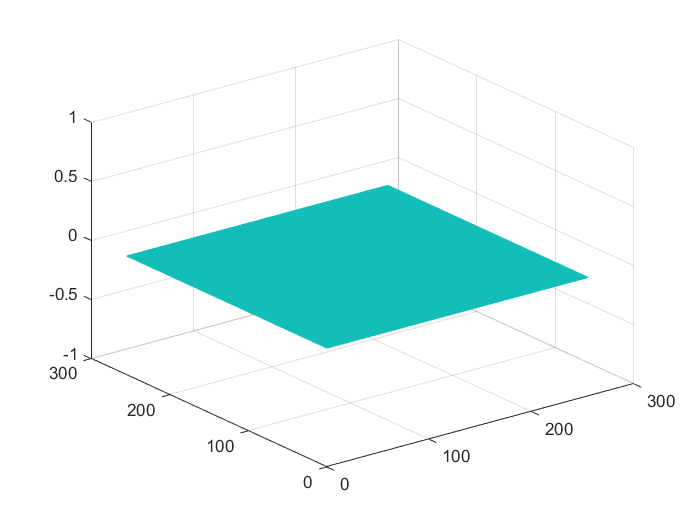} & \includegraphics[width=0.115\textwidth]{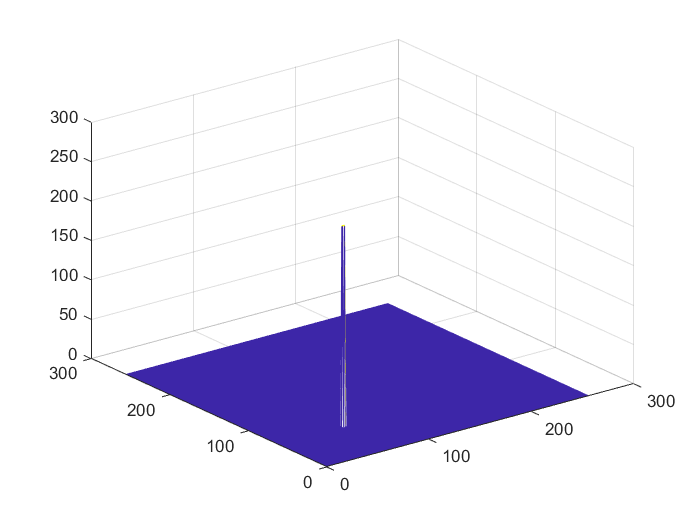} \\
    
        \includegraphics[width=0.115\textwidth]{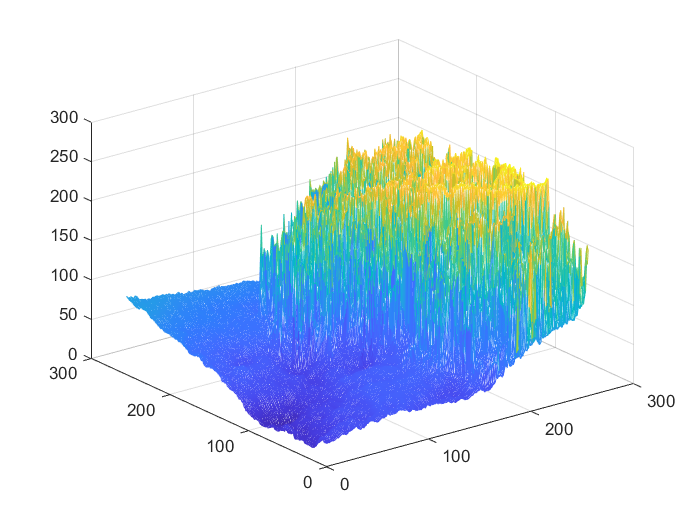} & \includegraphics[width=0.115\textwidth]{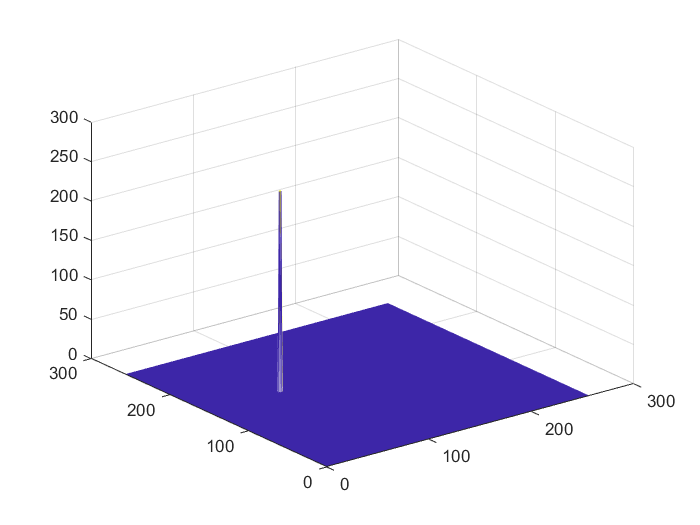} & \includegraphics[width=0.115\textwidth]{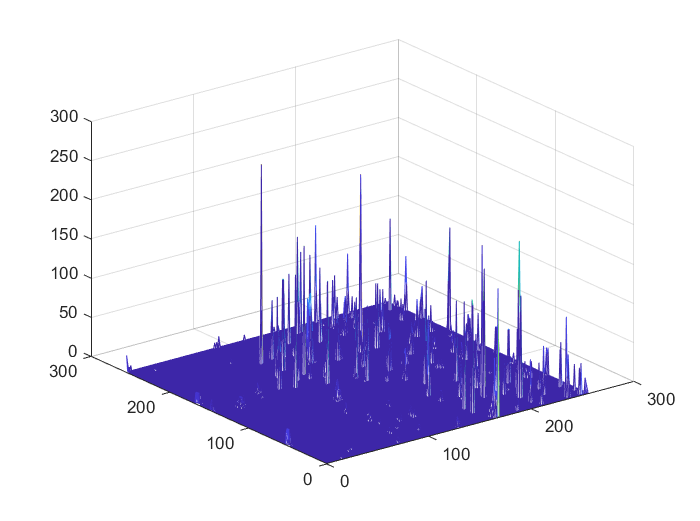} & \includegraphics[width=0.115\textwidth]{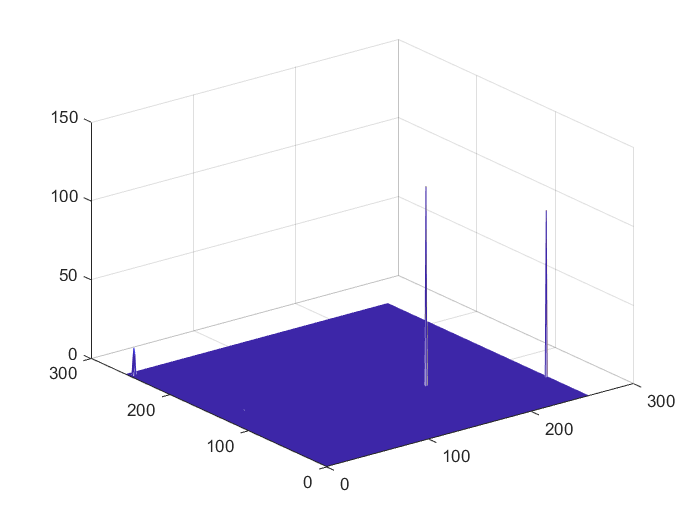} & \includegraphics[width=0.115\textwidth]{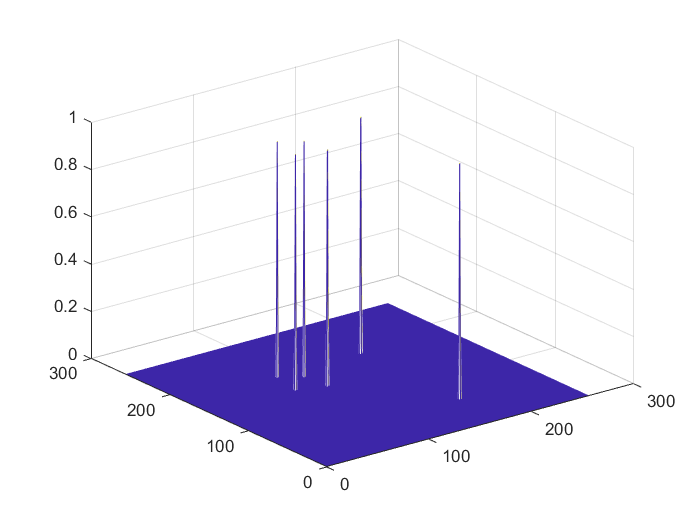} &  \includegraphics[width=0.115\textwidth]{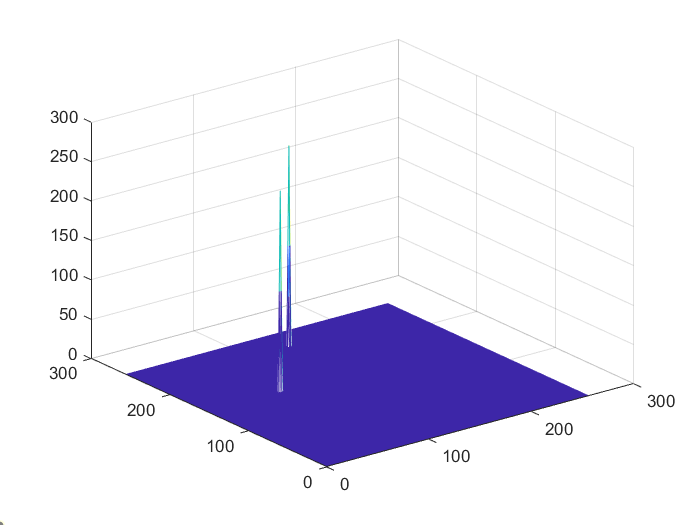} & \includegraphics[width=0.115\textwidth]{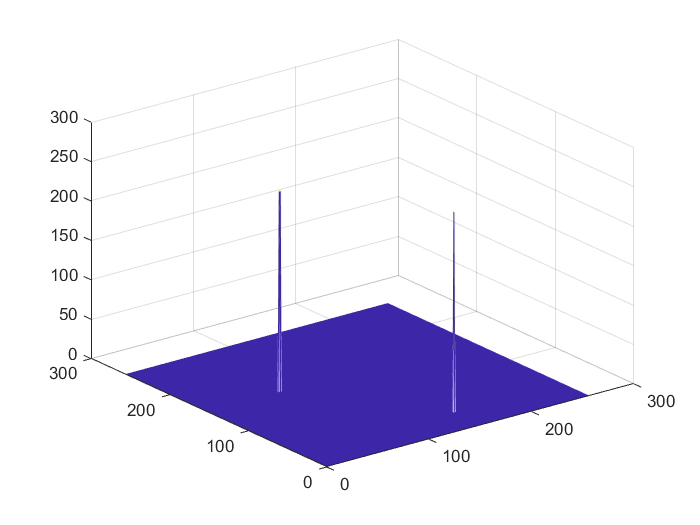} & \includegraphics[width=0.115\textwidth]{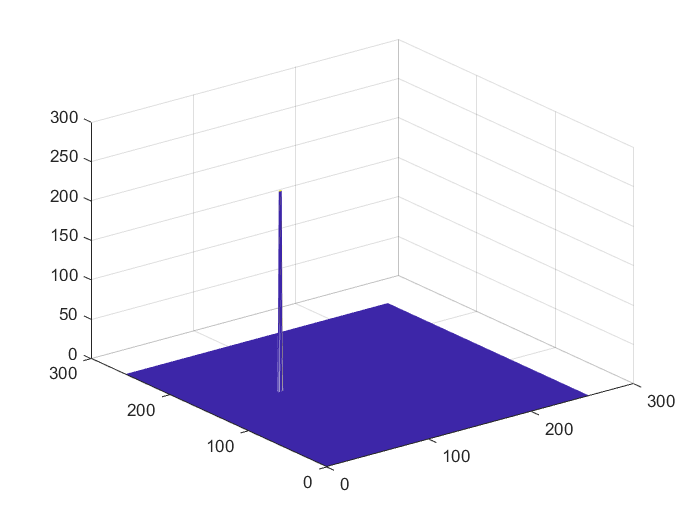} \\

        \includegraphics[width=0.115\textwidth]{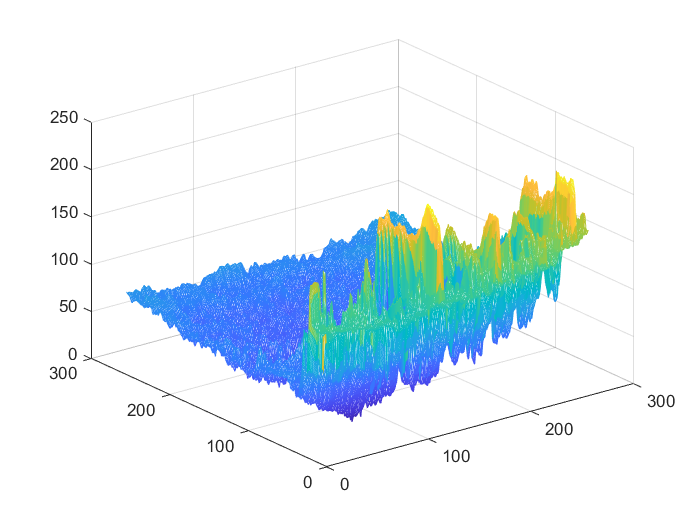} & \includegraphics[width=0.115\textwidth]{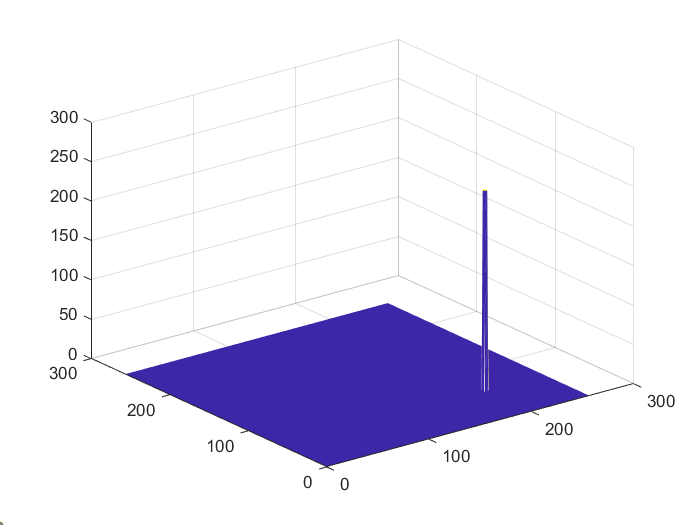} & \includegraphics[width=0.115\textwidth]{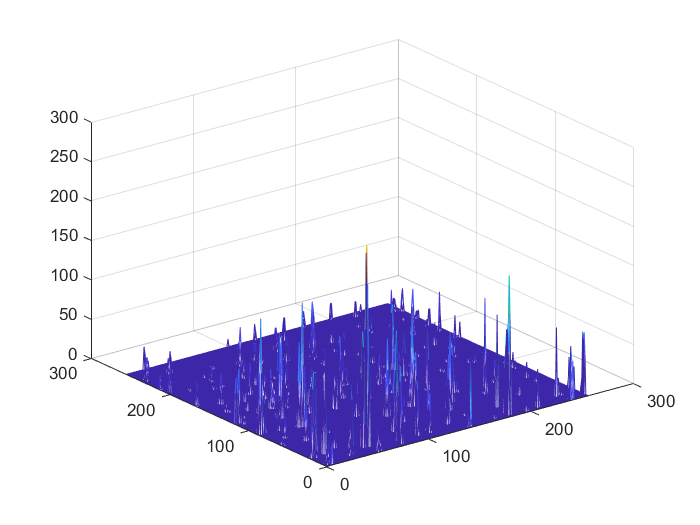} & \includegraphics[width=0.115\textwidth]{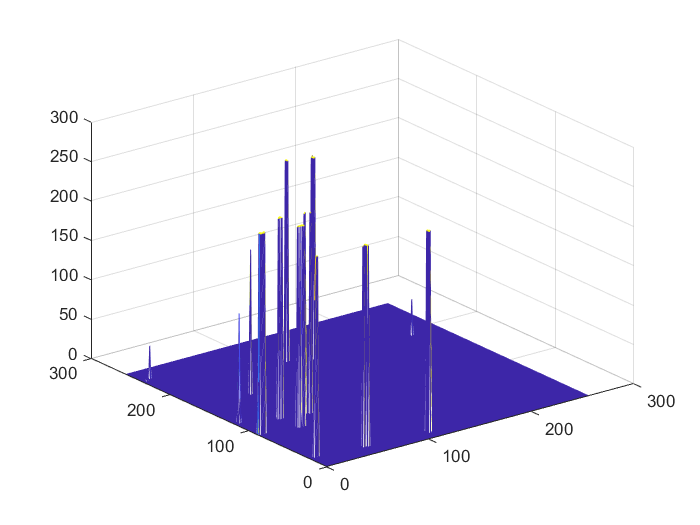} & \includegraphics[width=0.115\textwidth]{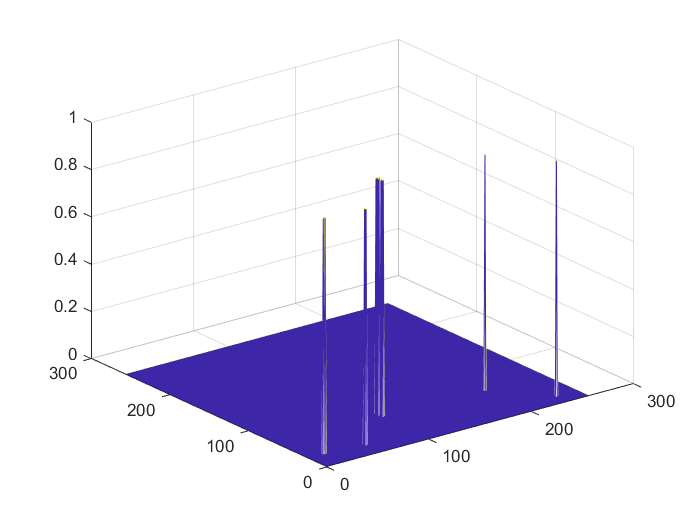} &  \includegraphics[width=0.115\textwidth]{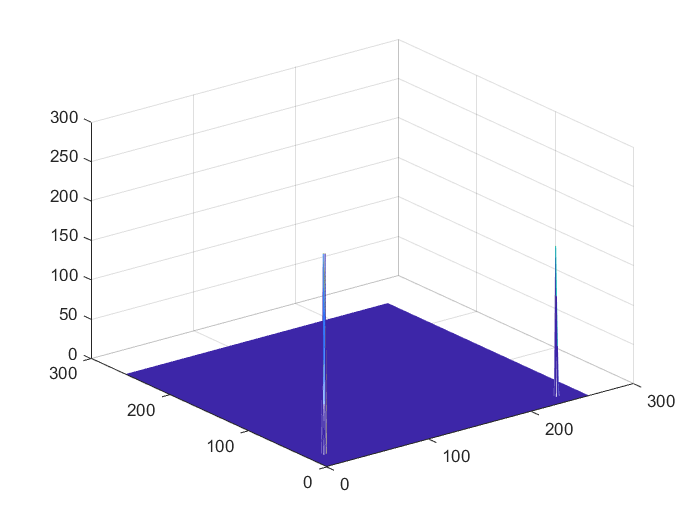} & \includegraphics[width=0.115\textwidth]{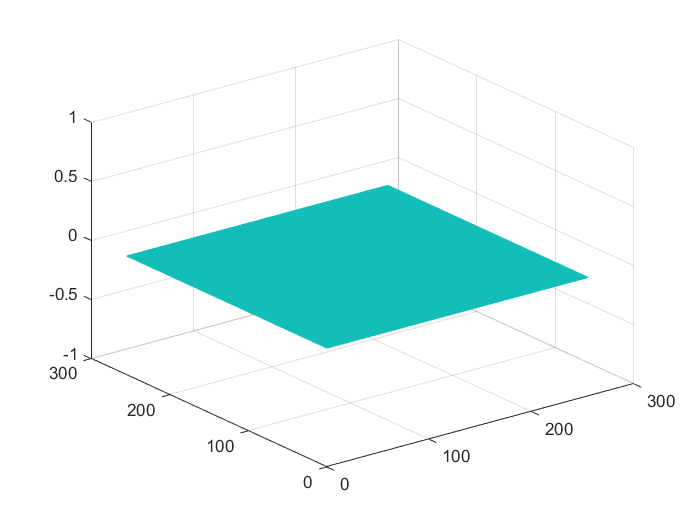} & \includegraphics[width=0.115\textwidth]{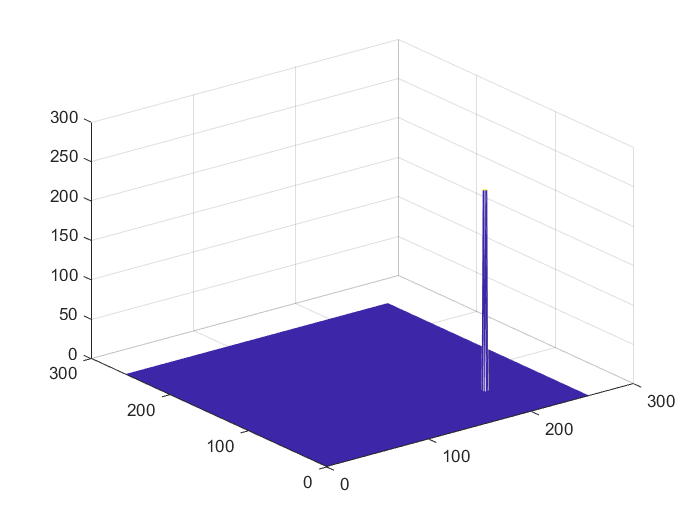} \\

    \end{tabular}
 \caption{3D visualization results corresponding to those in \cref{fig:results_visualization}. Different colors represent distinct grayscale values, ranging from blue to yellow, with values gradually increasing. Please zoom in for better viewing.}
     \label{fig:results_3dvisual}
\end{figure*}

\paragraph{Quantitative Results.}
The quantitative results of different methods are presented in \cref{tab:main_quantitative_result}. Overall, our MSHNet performs the best on all metrics and all datasets. 

As expected, traditional methods perform poorly on these challenging datasets due to the limitations of manually designed priors for feature extraction. In contrast, DL-based methods can automatically learn useful features and yield better results than traditional methods. Nevertheless, existing DL-based methods lack the consideration of different scales and locations of targets in the training stage. The predictions of existing DL-based methods suffer from the incomplete shape and missed detection of valid pixels,  \eg, lower IoU and ${\rm P_d}$.  
Compared with DNANet, which is one of the existing SOTA DL-based methods, our MSHNet achieves 1.45\% and 0.57\% higher IoU, 2.04\% and 1.06\% higher ${\rm P_d}$ on IRSTD-1k and NUAA-SIRST datasets, respectively. In addition, the false alarm (${\rm F_a}$) on both datasets are all reduced by a large margin, demonstrating the superiority of our method.

As mentioned before, there are no complex model structures in MSHNet, which achieves a better balance between detection performance, floating point of operations (FLOPs) and inference time consumption. The results of DL-based methods are presented in \cref{fig:motivation_t_iou}. It can be observed that MSHNet has the best detection performance and the least inference time consumption. It also has the least number of FLOPs except for ALCNet. 
We own the less inference time consumption of MSHNet to the smaller number of FLOPs and the simple architectures (complex structures in deep models bring more inference time consumption \cite{wang2023riformer}).
The better detection performance demonstrates the effectiveness of the proposed multi-scale head and the scale and location sensitive loss.

\paragraph{Qualitative Results.}
For different kinds of methods, we visualize the detection results of the representative methods in \cref{fig:results_visualization} and \cref{fig:results_3dvisual}. It can be observed that our method has better capability for the detection of challenging small targets. Traditional methods are susceptible to noise interference, resulting in a large number of false alarms. Moreover, due to the small scale and low contrast of the target, traditional methods and other DL-based methods struggle to effectively extract feature information of small targets from challenging scenarios, resulting in a considerable number of missed targets. In contrast, our methods can accurately distinguish small targets from these low-contrast infrared images. Take the case in the 5-th row in \cref{fig:results_visualization}  for example, except for MSHNet and WLSLCM, other methods fail to detect the target. Moreover, from the peaks and the area under the peaks in \cref{fig:results_3dvisual}, it can be observed that our method predicts targets with higher confidence and obtains a closer number of peaks with ground-truth. We attribute the superiority of MSHNet to the scale and location sensitivity brought by our SLS loss and multi-scale head.

\subsection{Discussions}
\label{sec:ablation_study}
We provid some discussions to show the effectiveness of our method.
Experiments are conducted on IRSTD-1k dataset.

\paragraph{Detection Results for Different Scales of Targets.}
Since our SLS loss and MSHNet both are scale sensitive, we show how the detection performance changes with respect to different scales of targets. \revise{To do this, we split the targets into three different scale levels according to the number of pixels: (0,10], (10, 40] and (40, $\infty$]. Results are shown in \cref{tab:aresults_for_different_scales}. } Due to the limited space, only several DL-based methods are evaluated (specifically, ISNet, DNANet and our MSHNet).
Compared with DNANet, MSHNet performs better for all \revise{scale levels } of targets. Compared with ISNet, MSHNet achieves comparable performance with ISNet for \revise{(0,10] and (10,40] scales } but much better performance for \revise{(40,$\infty$] scale}. The conclusion is that MSHNet can pay attention to different \revise{scale levels } of targets, resulting in an overall better performance.

\paragraph{Impact of Scale and Location Sensitive Loss.}
We compare our SLS loss with the commonly used IoU loss and Dice loss.
The results of several DL-based methods trained with different losses are presented in \cref{tab:ablation_multiloss}. 
Overall, the detection performance of different detectors is all improved by our SLS loss in terms of IoU metric, demonstrating the effectiveness and generalization of SLS loss. However, it is hard to tell which loss performs the best on the metric of ${\rm P_d}$. Interestingly, adopting our SLS loss results in a poorer performance on ${\rm F_a}$ metric. The reason is that more false alarm pixels may be treated as positive targets since our SLS loss tries to distinguish the targets of all different scales.

\begin{table}[]
\begin{center}
\setlength{\tabcolsep}{5pt}
\small
\begin{tabular}{c|c|c|c|c}
\hline

Scale &Method &IoU$\uparrow$ & ${\rm P_d}\uparrow$ & ${\rm F_a}\downarrow$ \\

\hline
\multirow{3}{*}{\makecell{\revise{(0,10]}}} 
& DNANet \cite{acmnet} &47.26 & 91.27 &22.53 \\
& ISNet \cite{dnanet} & \textbf{50.94} & 94.53 & 30.91 \\
& MSHNet (Ours) & 49.51 &\textbf{95.24} &\textbf{20.68} \\
\hline
\multirow{3}{*}{\makecell{\revise{(10,40]}}} 
& DNANet \cite{acmnet} &63.14 &91.85 &13.25 \\
& ISNet \cite{dnanet} &\textbf{65.41} &\textbf{91.91} &16.60 \\
& MSHNet (Ours) &64.79 &91.85 &\textbf{8.03} \\
\hline
\multirow{3}{*}{\makecell{\revise{(40,$\infty$]}}} 
& DNANet \cite{acmnet} &78.46 &93.94 &\textbf{6.78} \\
& ISNet \cite{dnanet} &66.45 &87.87 &58.80 \\
& MSHNet (Ours) &\textbf{79.20} &\textbf{96.97} &11.02 \\
\hline
\end{tabular}
\caption{\revise{Comparisons of different methods in terms of different scales of targets. Results for the metrics of IoU(\%), ${\rm P_d}$(\%) and ${\rm F_a}$($10^{-6}$) are presented. Our MSHNet achieves overall the best performance.}}
\label{tab:aresults_for_different_scales}
\end{center}
\vspace{-15pt}
\end{table}

\begin{table*}[]
\begin{center}
\setlength{\tabcolsep}{13pt}
\small
\begin{tabular}{c|c|c|c|c|c|c|c|c|c}
\hline
\multirow{2}{*}{Loss} & \multicolumn{3}{c|}{DNANet \cite{acmnet}} & \multicolumn{3}{c|}{ISNet \cite{dnanet}} & \multicolumn{3}{c}{MSHNet (Ours)} \\ \cline{2-10}
 & IoU$\uparrow$ & ${\rm P_d}\uparrow$ & ${\rm F_a}\downarrow$ & IoU$\uparrow$ & ${\rm P_d}\uparrow$ & ${\rm F_a}\downarrow$ & IoU$\uparrow$ & ${\rm P_d}\uparrow$ & ${\rm F_a}\downarrow$ \\ \hline
 $\mathcal{L}_{IoU}$ & 65.71 & 91.84 & 17.61 & 62.88 & \textbf{92.59} & 27.92 & 64.83 & 91.16 & \textbf{5.28} \\ \hline
 $\mathcal{L}_{Dice}$ & 65.58 & \textbf{92.17} & \textbf{12.75} & 62.94 & 90.91 & 19.26 & 65.36 & 92.18 & 14.12\\ \hline
 $\mathcal{L}_{SLS}$ (Ours) & \textbf{67.09} & 92.15 & 35.15 & \textbf{64.42} & 92.26 & \textbf{15.14} & \textbf{67.16} & \textbf{93.88} & 15.03 \\ \hline
\end{tabular}
\caption{\revise{Comparisons of different losses and detectors. Results for the metrics of IoU(\%), ${\rm P_d}$(\%) and ${\rm F_a}$($10^{-6}$) are presented. The main IoU metrics of different detectors are consistently boosted by our scale and location sensitive loss. }}
\label{tab:ablation_multiloss}
\end{center}
\vspace{-10pt}
\end{table*}

\begin{table}[]
\begin{center}
\setlength\tabcolsep{16pt}
\small
\begin{tabular}{c|c|c|c}
\hline
Loss & IoU$\uparrow$ & ${\rm P_d}\uparrow$ & ${\rm F_a}\downarrow$ \\ \hline
$\mathcal{L}_{S}$ w/o. $w$  & 64.83 & 91.16 & 5.28    \\ \hline
$\mathcal{L}_S$ & 65.82 & 89.46  & \textbf{4.06} \\ \hline
$\mathcal{L}_{S}$ + $\mathcal{L}_{L}$ & \textbf{67.16} & \textbf{93.88} & 15.03 \\ \hline
\end{tabular}
\caption{\revise{Ablation study of the scale sensitive loss ($\mathcal{L}_{S}$) and location sensitive loss ($\mathcal{L}_{L}$). Results are obtained by MSHNet. The metrics of IoU(\%), ${\rm P_d}$(\%) and ${\rm F_a}$ ($10^{-6}$) are adopted. Note that $\mathcal{L}_{S}$ without $w$ is equivalent to $\mathcal{L}_{IoU}$.}}
\label{tab:ablation_SLSloss}
\end{center}
\vspace{-5pt}
\end{table}

\begin{table}[]
\begin{center}
\setlength\tabcolsep{17pt}
\small
\begin{tabular}{c|c|c|c}
\hline
Loss & IoU$\uparrow$ & ${\rm P_d}\uparrow$ & ${\rm F_a}\downarrow$ \\ \hline
$\mathcal{L}_{S}$ + $\mathcal{L}_{L_2}$  & 64.89 & 92.86 & 7.29 \\ \hline
\revise{$\mathcal{L}_{S}$ + $\mathcal{L}_{L_1}$}  & \revise{64.75} & \revise{93.54} & \revise{\textbf{6.18}} \\ \hline
$\mathcal{L}_{S}$ + $\mathcal{L}_{L}$ & \textbf{67.16} & \textbf{93.88} & 15.03 \\ \hline

\end{tabular}
\caption{\revise{Ablation study of different types of location sensitive loss. Results are obtained by MSHNet. The metrics of IoU(\%), ${\rm P_d}$(\%) and ${\rm F_a}$ ($10^{-6}$) are adopted.}}
\label{tab:ablation_lloss}
\end{center}
\vspace{-10pt}
\end{table}

Next, we show the effectiveness of scale sensitive loss ($\mathcal{L}_{S}$) and location sensitive loss ($\mathcal{L}_{L}$) separately. To do this, we train MSHNet with different losses. The results are presented in \cref{tab:ablation_SLSloss}. Note that the difference between $\mathcal{L}_{S}$ and $\mathcal{L}_{IoU}$ lies in the newly introduced weight $w$ (refer to \cref{sec:scale_sensitive_loss}). By comparing the results in the first two rows, we can find that the introduction of $w$ (refer to \cref{eq:loss_s}) is positive to the metrics of IoU and ${\rm F_a}$, indicating that the detector produces more accurate shapes for targets as well as less false alarms.  While comparing the results in the last two rows, it can be observed that more false alarms will be produced when the location sensitive loss is introduced. However, both IoU and ${\rm P_d}$ are greatly improved.

Finally, we make a study on different types of location sensitive loss. As mentioned in \cref{sec:location_loss}, our location sensitive loss can distinguish different location errors \revise{more effectively }. To show the effectiveness of such distinguishability, we use the $L_2$ \revise{and $L_1$ distances } between $\mathbf{c}_p$ and $\mathbf{c}_{gt}$ as the location sensitive loss (denoted as $\mathcal{L}_{L_2}$ \revise{and $\mathcal{L}_{L_1}$, respectively}). Results are shown in \cref{tab:ablation_lloss}. As we can see, the distinguishability of different location errors is effective in getting better IoU and ${\rm P_d}$ metrics. However, more false alarms are produced \revise{(higher $\rm F_a$)}. Such phenomenon is also observed from \cref{tab:ablation_SLSloss}. The reason may be that the distinguishability of different location errors makes the detector more sensitive to small targets, resulting in the potential of treating some noises to targets.

\paragraph{Impact of Multi-Scale Heads.}
Now we ablate the number of scales in our multi-scale head. Results are shown in \cref{tab:ablation_head}. 
As we can see, different predictions achieve different detection performance. 
 Overall, the more scales adopted, the better the detection performance is achieved. For example, the IoU is improved from \revise{63.10\% }  to 67.16\% when the number of scales is increased from 1 to 4. We use 4 scales by default.

\begin{table}[]
\begin{center}
\setlength\tabcolsep{4pt}
\small
\begin{tabular}{c|c|c|c|c}
\hline
\makecell{Num. of Scales} &\makecell{\revise{Supervised Predictions}} &IoU$\uparrow$ & ${\rm P_d}\uparrow$ & ${\rm F_a}\downarrow$ \\ \hline
 \revise{1} & \revise{$p$} & \revise{63.10} &\revise{86.73} &\revise{19.21}\\ \hline
 1 & \revise{$p_4, p$} & 64.69 & 93.87 & 36.74 \\ \hline
 2 & \revise{$p_3,p_4, p$} & 64.35 & \textbf{94.89} & 35.38 \\ \hline
 3 & \revise{$p_2, p_3, p_4, p$} & 65.90 & 93.54 & 24.37 \\ \hline
 4 & \revise{$p_1, p_2, p_3, p_4, p$} &  \textbf{67.16} & 93.88 & \textbf{15.03} \\ \hline
\end{tabular}
\caption{\revise{Ablation study of the number of scales in MSHNet. While reducing the number of adopted scales, the remaining smallest scale is removed. Results for the metrics of IoU(\%), ${\rm P_d}$(\%) and ${\rm F_a}$ ($10^{-6}$) are presented.}}
\label{tab:ablation_head}
\end{center}
\vspace{-10pt}
\end{table}

\section{Conclusion}
\label{sec:concles lusion}
In this paper, we focus on boosting the performance of infrared small target detection with an effective loss function but a simpler model structure. To do this, a Scale and Location Sensitive (SLS) loss is first proposed. The merits of SLS loss include: (1) it pays more attention to the targets that have large gaps between the predicted and ground-truth scales; (2) it distinguishes different types of location errors between the predicted and ground-truth center points of targets. 
Then a simple Multi-Scale Head is introduced to the plain U-Net (MSHNet), which produces multi-scale predictions for each input. Through applying SLS loss to different scales of predictions, different scales of targets can attract different attention from the detector, resulting in an overall better detection performance. Experimental results show that MSHNet achieves SOTA detection performance \revise{with the better balance of inference time and the number of floating point of operations}. While applying our SLS loss to other existing detectors, the overall detection performance can be boosted. 
However, more false alarms may be introduced. Through the analysis in \cref{sec:ablation_study}, we find that such a phenomenon is brought by the location sensitive loss in SLS loss \revise{which potentially treats some noises as targets}. In future works, we try to handle this by designing more suitable location sensitive loss.

\vspace{10pt}
\noindent\textbf{Acknowledgements}

\revise{\noindent This work was supported by the National Natural Science Foundation of China (62331006,62171038, 62088101, and 62371175), the R\&D Program of Beijing Municipal Education Commission (KZ202211417048), the Key R\&D Program of Zhejiang (2023C01044), the Fundamental Research Funds for the Provincial Universities of Zhejiang (GK239909299001-013), and the Fundamental Research Funds for the Central Universities.}

{
    \small
    \bibliographystyle{ieeenat_fullname}
    \bibliography{main}
}

\end{document}

%% file: preamble.tex
\usepackage[dvipsnames]{xcolor}


%% file: main.bbl
\begin{thebibliography}{35}
\providecommand{\natexlab}[1]{#1}
\providecommand{\url}[1]{\texttt{#1}}
\expandafter\ifx\csname urlstyle\endcsname\relax
  \providecommand{\doi}[1]{doi: #1}\else
  \providecommand{\doi}{doi: \begingroup \urlstyle{rm}\Url}\fi

\bibitem[Chen et~al.(2023)Chen, Fu, Wei, Zheng, and Heide]{chen2023instance}
Linwei Chen, Ying Fu, Kaixuan Wei, Dezhi Zheng, and Felix Heide.
\newblock Instance segmentation in the dark.
\newblock \emph{International Journal of Computer Vision}, 131\penalty0 (8):\penalty0 2198--2218, 2023.

\bibitem[Dai and Wu(2017)]{RIPT}
Yimian Dai and Yiquan Wu.
\newblock Reweighted infrared patch-tensor model with both nonlocal and local priors for single-frame small target detection.
\newblock \emph{IEEE Journal of Selected Topics in Applied Earth Observations and Remote Sensing}, 10\penalty0 (8):\penalty0 3752--3767, 2017.

\bibitem[Dai et~al.(2021{\natexlab{a}})Dai, Wu, Zhou, and Barnard]{acmnet}
Yimian Dai, Yiquan Wu, Fei Zhou, and Kobus Barnard.
\newblock Asymmetric contextual modulation for infrared small target detection.
\newblock In \emph{Proceedings of the IEEE/CVF Winter Conference on Applications of Computer Vision}, pages 950--959, 2021{\natexlab{a}}.

\bibitem[Dai et~al.(2021{\natexlab{b}})Dai, Wu, Zhou, and Barnard]{alcnet}
Yimian Dai, Yiquan Wu, Fei Zhou, and Kobus Barnard.
\newblock Attentional local contrast networks for infrared small target detection.
\newblock \emph{IEEE Transactions on Geoscience and Remote Sensing}, 59\penalty0 (11):\penalty0 9813--9824, 2021{\natexlab{b}}.

\bibitem[Deshpande et~al.(1999)Deshpande, Er, Venkateswarlu, and Chan]{max_mean}
Suyog~D. Deshpande, Meng~Hwa Er, Ronda Venkateswarlu, and Philip Chan.
\newblock {Max-mean and max-median filters for detection of small targets}.
\newblock In \emph{Signal and Data Processing of Small Targets 1999}, pages 74 -- 83. International Society for Optics and Photonics, SPIE, 1999.

\bibitem[Fu et~al.(2020)Fu, Zheng, Zhang, Zheng, and Huang]{fu2020simultaneous}
Ying Fu, Yongrong Zheng, Lin Zhang, Yinqiang Zheng, and Hua Huang.
\newblock Simultaneous hyperspectral image super-resolution and geometric alignment with a hybrid camera system.
\newblock \emph{Neurocomputing}, 384:\penalty0 282--294, 2020.

\bibitem[Fu et~al.(2021)Fu, Chen, Zhang, and Lin]{fuying-2021-neurocomputing}
Ying Fu, Jian Chen, Tao Zhang, and Yonggang Lin.
\newblock Residual scale attention network for arbitrary scale image super-resolution.
\newblock \emph{Neurocomputing}, 427:\penalty0 201--211, 2021.

\bibitem[Gao et~al.(2013)Gao, Meng, Yang, Wang, Zhou, and Hauptmann]{ipi}
Chenqiang Gao, Deyu Meng, Yi Yang, Yongtao Wang, Xiaofang Zhou, and Alexander~G Hauptmann.
\newblock Infrared patch-image model for small target detection in a single image.
\newblock \emph{IEEE Transactions on Image Processing}, 22\penalty0 (12):\penalty0 4996--5009, 2013.

\bibitem[Han et~al.(2019)Han, Moradi, Faramarzi, Liu, Zhang, and Zhao]{tllcm}
Jinhui Han, Saed Moradi, Iman Faramarzi, Chengyin Liu, Honghui Zhang, and Qian Zhao.
\newblock A local contrast method for infrared small-target detection utilizing a tri-layer window.
\newblock \emph{IEEE Geoscience and Remote Sensing Letters}, 17\penalty0 (10):\penalty0 1822--1826, 2019.

\bibitem[Han et~al.(2020)Han, Moradi, Faramarzi, Zhang, Zhao, Zhang, and Li]{wslcm}
Jinhui Han, Saed Moradi, Iman Faramarzi, Honghui Zhang, Qian Zhao, Xiaojian Zhang, and Nan Li.
\newblock Infrared small target detection based on the weighted strengthened local contrast measure.
\newblock \emph{IEEE Geoscience and Remote Sensing Letters}, 18\penalty0 (9):\penalty0 1670--1674, 2020.

\bibitem[Hou et~al.(2021)Hou, Wang, Tan, Zhao, Zheng, and Zhang]{likelihood_loss}
Qingyu Hou, Zhipeng Wang, Fanjiao Tan, Ye Zhao, Haoliang Zheng, and Wei Zhang.
\newblock Ristdnet: Robust infrared small target detection network.
\newblock \emph{IEEE Geoscience and Remote Sensing Letters}, 19:\penalty0 1--5, 2021.

\bibitem[Huang et~al.(2019)Huang, Nie, Zheng, and Fu]{huang2019image}
Hua Huang, Guangtao Nie, Yinqiang Zheng, and Ying Fu.
\newblock Image restoration from patch-based compressed sensing measurement.
\newblock \emph{Neurocomputing}, 340:\penalty0 145--157, 2019.

\bibitem[Ju et~al.(2021)Ju, Luo, Liu, and Luo]{irstd_giou}
Moran Ju, Jiangning Luo, Guangqi Liu, and Haibo Luo.
\newblock Istdet: An efficient end-to-end neural network for infrared small target detection.
\newblock \emph{Infrared Physics \& Technology}, 114:\penalty0 103659, 2021.

\bibitem[Li et~al.(2022)Li, Xiao, Wang, Wang, Lin, Li, An, and Guo]{dnanet}
Boyang Li, Chao Xiao, Longguang Wang, Yingqian Wang, Zaiping Lin, Miao Li, Wei An, and Yulan Guo.
\newblock Dense nested attention network for infrared small target detection.
\newblock \emph{IEEE Transactions on Image Processing}, 32:\penalty0 1745--1758, 2022.

\bibitem[Liu et~al.(2017)Liu, Du, Zhao, Dong, Hui, and Wang]{liuinfrared}
Ming Liu, Hao-yuan Du, Yue-jin Zhao, Li-quan Dong, Mei Hui, and SX Wang.
\newblock Image small target detection based on deep learning with snr controlled sample generation.
\newblock \emph{Current Trends in Computer Science and Mechanical Automation}, 1:\penalty0 211--220, 2017.

\bibitem[Rezatofighi et~al.(2019)Rezatofighi, Tsoi, Gwak, Sadeghian, Reid, and Savarese]{giou}
Hamid Rezatofighi, Nathan Tsoi, JunYoung Gwak, Amir Sadeghian, Ian Reid, and Silvio Savarese.
\newblock Generalized intersection over union: A metric and a loss for bounding box regression.
\newblock In \emph{Proceedings of the IEEE/CVF Conference on Computer Vision and Pattern Recognition}, pages 658--666, 2019.

\bibitem[Rivest and Fortin(1996)]{tophat}
Jean-Francois Rivest and Roger Fortin.
\newblock {Detection of dim targets in digital infrared imagery by morphological image processing}.
\newblock \emph{Optical Engineering}, 35\penalty0 (7):\penalty0 1886 -- 1893, 1996.

\bibitem[Sudre et~al.(2017)Sudre, Li, Vercauteren, Ourselin, and Jorge~Cardoso]{diceloss}
Carole~H Sudre, Wenqi Li, Tom Vercauteren, Sebastien Ourselin, and M Jorge~Cardoso.
\newblock Generalised dice overlap as a deep learning loss function for highly unbalanced segmentations.
\newblock In \emph{Deep Learning in Medical Image Analysis and Multimodal Learning for Clinical Decision}, pages 240--248. Springer, 2017.

\bibitem[Sun et~al.(2020)Sun, Yang, and An]{MSLSTIPT}
Yang Sun, Jungang Yang, and Wei An.
\newblock Infrared dim and small target detection via multiple subspace learning and spatial-temporal patch-tensor model.
\newblock \emph{IEEE Transactions on Geoscience and Remote Sensing}, 59\penalty0 (5):\penalty0 3737--3752, 2020.

\bibitem[Teutsch and Kr{\"u}ger(2010)]{maritime_1}
Michael Teutsch and Wolfgang Kr{\"u}ger.
\newblock Classification of small boats in infrared images for maritime surveillance.
\newblock In \emph{International WaterSide Security Conference}, pages 1--7. IEEE, 2010.

\bibitem[Wang et~al.(2019)Wang, Zhou, and Wang]{mdvsfa}
Huan Wang, Luping Zhou, and Lei Wang.
\newblock Miss detection vs. false alarm: Adversarial learning for small object segmentation in infrared images.
\newblock In \emph{Proceedings of the IEEE/CVF International Conference on Computer Vision}, pages 8509--8518, 2019.

\bibitem[Wang et~al.(2023)Wang, Zhang, Liu, Wu, Yang, Liu, Chen, Luo, and Lin]{wang2023riformer}
Jiahao Wang, Songyang Zhang, Yong Liu, Taiqiang Wu, Yujiu Yang, Xihui Liu, Kai Chen, Ping Luo, and Dahua Lin.
\newblock Riformer: Keep your vision backbone effective but removing token mixer.
\newblock In \emph{Proceedings of the IEEE/CVF Conference on Computer Vision and Pattern Recognition}, pages 14443--14452, 2023.

\bibitem[Wang et~al.(2022)Wang, Du, Liu, and Cao]{irstd_ciou}
Kewei Wang, Shuaiyuan Du, Chengxin Liu, and Zhiguo Cao.
\newblock Interior attention-aware network for infrared small target detection.
\newblock \emph{IEEE Transactions on Geoscience and Remote Sensing}, 60:\penalty0 1--13, 2022.

\bibitem[Wei et~al.(2022)Wei, Avil{\'e}s-Rivero, Liang, Fu, Huang, and Sch{\"o}nlieb]{wei2022tfpnp}
Kaixuan Wei, Angelica~I Avil{\'e}s-Rivero, Jingwei Liang, Ying Fu, Hua Huang, and Carola-Bibiane Sch{\"o}nlieb.
\newblock Tfpnp: Tuning-free plug-and-play proximal algorithms with applications to inverse imaging problems.
\newblock \emph{J. Mach. Learn. Res.}, 23\penalty0 (16):\penalty0 1--48, 2022.

\bibitem[Wu et~al.(2022)Wu, Hong, and Chanussot]{uiunet}
Xin Wu, Danfeng Hong, and Jocelyn Chanussot.
\newblock Uiu-net: U-net in u-net for infrared small object detection.
\newblock \emph{IEEE Transactions on Image Processing}, 32:\penalty0 364--376, 2022.

\bibitem[Ying et~al.(2023)Ying, Liu, Wang, Li, Chen, Lin, Sheng, and Zhou]{ying2023mapping}
Xinyi Ying, Li Liu, Yingqian Wang, Ruojing Li, Nuo Chen, Zaiping Lin, Weidong Sheng, and Shilin Zhou.
\newblock Mapping degeneration meets label evolution: Learning infrared small target detection with single point supervision.
\newblock In \emph{Proceedings of the IEEE/CVF Conference on Computer Vision and Pattern Recognition}, pages 15528--15538, 2023.

\bibitem[Zhang et~al.(2021{\natexlab{a}})Zhang, Ni, Yan, and Zhang]{field_1}
Ke Zhang, Shuyan Ni, Dashuang Yan, and Aidi Zhang.
\newblock Review of dim small target detection algorithms in single-frame infrared images.
\newblock In \emph{IEEE Advanced Information Management, Communicates, Electronic and Automation Control Conference}, pages 2115--2120, 2021{\natexlab{a}}.

\bibitem[Zhang and Peng(2019)]{PSTNN}
Landan Zhang and Zhenming Peng.
\newblock Infrared small target detection based on partial sum of the tensor nuclear norm.
\newblock \emph{Remote Sensing}, 11\penalty0 (4):\penalty0 382, 2019.

\bibitem[Zhang et~al.(2018)Zhang, Peng, Zhang, Cao, and Peng]{nram}
Landan Zhang, Lingbing Peng, Tianfang Zhang, Siying Cao, and Zhenming Peng.
\newblock Infrared small target detection via non-convex rank approximation minimization joint $l_{2, 1}$ norm.
\newblock \emph{Remote Sensing}, 10\penalty0 (11):\penalty0 1821, 2018.

\bibitem[Zhang et~al.(2022{\natexlab{a}})Zhang, Zhang, Yang, Bai, Zhang, and Guo]{zhang2022isnet}
Mingjin Zhang, Rui Zhang, Yuxiang Yang, Haichen Bai, Jing Zhang, and Jie Guo.
\newblock Isnet: Shape matters for infrared small target detection.
\newblock In \emph{Proceedings of the IEEE/CVF Conference on Computer Vision and Pattern Recognition}, pages 877--886, 2022{\natexlab{a}}.

\bibitem[Zhang et~al.(2021{\natexlab{b}})Zhang, Fu, and Li]{zhang2021hyperspectral}
Tao Zhang, Ying Fu, and Cheng Li.
\newblock Hyperspectral image denoising with realistic data.
\newblock In \emph{Proceedings of the IEEE/CVF International Conference on Computer Vision}, pages 2248--2257, 2021{\natexlab{b}}.

\bibitem[Zhang et~al.(2022{\natexlab{b}})Zhang, Fu, and Li]{zhang2022deep}
Tao Zhang, Ying Fu, and Cheng Li.
\newblock Deep spatial adaptive network for real image demosaicing.
\newblock In \emph{Proceedings of the AAAI Conference on Artificial Intelligence}, pages 3326--3334, 2022{\natexlab{b}}.

\bibitem[Zhang et~al.(2022{\natexlab{c}})Zhang, Fu, and Zhang]{zhang2022guided}
Tao Zhang, Ying Fu, and Jun Zhang.
\newblock Guided hyperspectral image denoising with realistic data.
\newblock \emph{International Journal of Computer Vision}, 130\penalty0 (11):\penalty0 2885--2901, 2022{\natexlab{c}}.

\bibitem[Zhao et~al.(2022)Zhao, Li, Li, Hu, Ma, and Tao]{field_2}
Mingjing Zhao, Wei Li, Lu Li, Jin Hu, Pengge Ma, and Ran Tao.
\newblock Single-frame infrared small-target detection: A survey.
\newblock \emph{IEEE Geoscience and Remote Sensing Magazine}, 10\penalty0 (2):\penalty0 87--119, 2022.

\bibitem[Zheng et~al.(2020)Zheng, Wang, Liu, Li, Ye, and Ren]{diou}
Zhaohui Zheng, Ping Wang, Wei Liu, Jinze Li, Rongguang Ye, and Dongwei Ren.
\newblock Distance-iou loss: Faster and better learning for bounding box regression.
\newblock In \emph{Proceedings of the AAAI Conference on Artificial Intelligence}, pages 12993--13000, 2020.

\end{thebibliography}
